\documentclass{article} %
\usepackage{bridgewam,times}

\usepackage{amsmath,amsfonts,bm}

\def\eqref#1{equation~\ref{#1}}

\def\1{\bm{1}}

\DeclareMathAlphabet{\mathsfit}{\encodingdefault}{\sfdefault}{m}{sl}
\SetMathAlphabet{\mathsfit}{bold}{\encodingdefault}{\sfdefault}{bx}{n}

\usepackage{graphicx}
\graphicspath{{section/photo/}}
\usepackage{hyperref}
\hypersetup{hypertexnames=false}
\usepackage{url}
\usepackage{graphicx}  %
\usepackage{booktabs}  %
\usepackage{makecell}  %
\usepackage[table]{xcolor}
\definecolor{bridgewamrow}{HTML}{C9E7DC}

\usepackage{pifont}
\usepackage{float}
\usepackage{tcolorbox}
\usepackage{helvet}
\definecolor{arxivcover}{HTML}{F5F8FC}
\definecolor{arxivcoverlink}{HTML}{285B48}

\makeatletter
\newenvironment{arxivfrontmatter}{%
  \begin{tcolorbox}[
    width=\textwidth,
    colback=arxivcover,
    colframe=arxivcover,
    boxrule=0pt,
    arc=6pt,
    boxsep=0pt,
    left=14pt,right=14pt,top=14pt,bottom=14pt,
    before skip=0pt,after skip=10pt,
    fontupper=\normalfont\normalsize
  ]%
  \setlength{\parindent}{0pt}%
  \setlength{\parskip}{0pt}%
  \hypersetup{hidelinks}%
  \renewcommand{\maketitle}{%
    {\sffamily\bfseries\fontsize{18}{21}\selectfont
     \raggedright\def\sysname{BridgeWAM}\@title\par}%
    \vspace{12pt}%
    \@author\par
    \vspace{14pt}%
  }%
  \renewenvironment{abstract}{%
    {\sffamily\bfseries\fontsize{10}{12}\selectfont Abstract\par}%
    \vspace{5pt}%
    \normalfont\fontsize{10}{11.5}\selectfont
  }{\par}%
}{\end{tcolorbox}}
\makeatother

\newcommand{\arxivcodelink}[2]{%
  \par\vspace{10pt}%
  {\sffamily\fontsize{9}{11}\selectfont
   \textbf{Code:}\enspace
   \href{#1}{\textcolor{arxivcoverlink}{\nolinkurl{#2}}}\par}%
}

\providecommand{\cmark}{\ding{51}}
\providecommand{\xmark}{\ding{55}}

\title{\sysname{}: Bridging Video and Action \\ Experts via Latent Bridge Queries}

\author{%
  {\sffamily\bfseries\fontsize{10}{12}\selectfont\raggedright
    Jiaxin Fang\textsuperscript{1,2}\enspace
    Ruitong Liu\textsuperscript{4}\enspace
    Zeyu Lai\textsuperscript{5,2}\enspace
    Xingyu Wang\textsuperscript{2,6}\enspace
    Mingyuan Jia\textsuperscript{2,3}\\[2pt]
    Pingyue Sheng\textsuperscript{2,3}\enspace
    Weikang Wan\textsuperscript{2,5}\enspace
    Mengdi Xu\textsuperscript{3,\textdagger}\enspace
    Yuxiang Gao\textsuperscript{2,\textdagger}\endgraf}
  \vspace{5pt}
  {\normalfont\fontsize{9}{11}\selectfont\raggedright
    \textsuperscript{1}Fudan University\enspace
    \textsuperscript{2}CocoMatrix\enspace
    \textsuperscript{3}Tsinghua University\enspace
    \textsuperscript{4}Peking University\\
    \textsuperscript{5}University of California, San Diego\enspace
    \textsuperscript{6}Institute of automation, Chinese academy of science\endgraf}%
}

\newcommand{\sysname}{{\textsc{BridgeWAM}}}
\newcommand{\queryname}{{\textsc{LBQs}}}

\begin{document}

\fancyhead{}
\renewcommand{\headrulewidth}{0pt}
\vspace*{-24pt}
\begin{arxivfrontmatter}
\maketitle

\begin{abstract}
World Action Models (WAMs) commonly couple video and action experts through dense, layer-wise Mixture-of-Transformers (MoT) attention, exposing action prediction to irrelevant visual features while enforcing matched expert depths. This coupling hinders selective visual-feature use and introduces redundant computation. We introduce Bridge-of-Experts (BoE), instantiated as \sysname{}, which aggregates visual, language, and proprioceptive priors through 32 Latent Bridge Queries (LBQs) into a shared, limited latent space. Joint video--action supervision lets action gradients guide the Video Expert toward representations better suited for action prediction, while the Action Expert learns to selectively retrieve task-relevant information through cross-attention. Decoupling expert depths and reusing LBQs during action denoising yield a $5.06$B model with a 30-layer Video Expert and a two-layer Action Expert, achieving $98.25\%$ success on LIBERO and $72.24\%$ pooled success on LIBERO-Plus, demonstrating effective task execution and strong generalization with an ultralight action head. Ablations support joint representation adaptation and unified LBQ conditioning. Relative to Fast-WAM, \sysname{} reduces single-sample inference FLOPs by $28.72\%$ in simulation profiling and real-robot latency to $77.1$\,ms ($2.47\times$ speedup). Qualitative experimental results further indicate preferential attention to target objects and task destinations across perturbations.
\end{abstract}
\arxivcodelink{https://github.com/jxfang04/BridgeWAM-code.git}{github.com/jxfang04/BridgeWAM-code}
\end{arxivfrontmatter}
\begingroup
\renewcommand{\thefootnote}{\fnsymbol{footnote}}
\footnotetext[2]{Corresponding authors.}
\endgroup

\begingroup
\setlength{\intextsep}{0pt}
\setlength{\parskip}{0pt}
\setlength{\abovecaptionskip}{-7pt}
\begin{figure}[H]
    \centering
    \includegraphics[width=\linewidth,trim={7bp 10bp 109bp 12bp},clip]{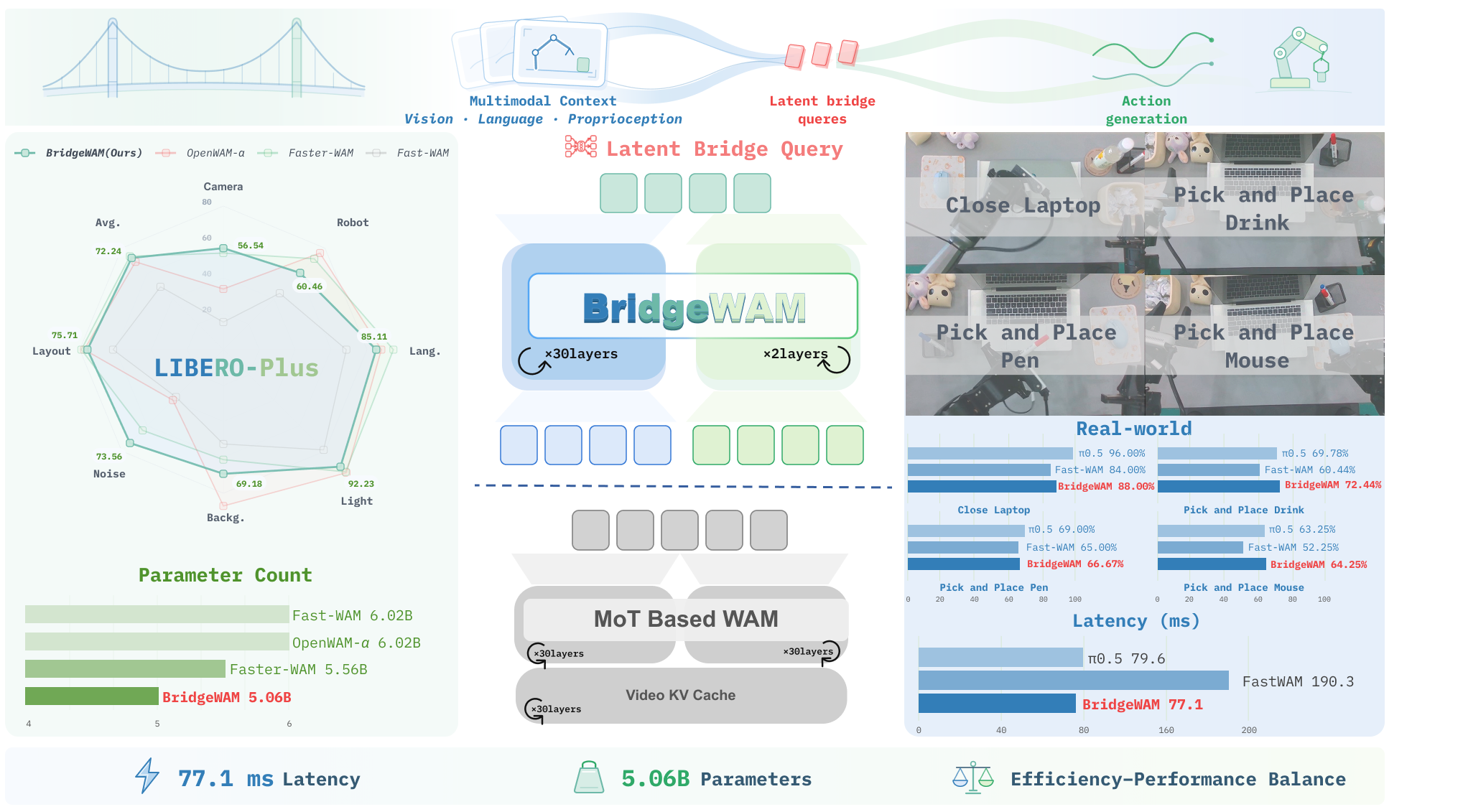}
    \caption{\textbf{Overview of \sysname{}.}
    Left: LIBERO-Plus success rates and parameters.
    Center: LBQs replace dense MoT video K/V, linking 30-layer Video and two-layer Action Experts.
    Right: real-world manipulation tasks, success rates, and measured inference latency in evaluated settings.}
    \label{fig:bridgewam_teaser}
\end{figure}
\endgroup

\section{Introduction}

World Action Models (WAMs) aim to predict robot actions from visual observations, language instructions, and proprioceptive state while retaining the broad visual priors of video-generation models. Useful WAMs must therefore simultaneously predict precise action chunks, remain robust when scenes and objects change, and run with low enough latency for closed-loop control. Recent WAMs have improved these properties by coupling a Video Expert and an Action Expert with Mixture-of-Transformers (MoT) attention~\cite{ye2026dreamzero,bi2025motus,li2026lingbotva,yuan2026fastwam}. However, information transfer between the two modalities remains largely unchanged.

WAM designs progressively simplify inference from joint video--action generation to lighter visual conditioning. Joint models process current and future video, action, and state tokens together~\cite{ye2026dreamzero}. IDM-style models use asymmetric attention: the video branch operates independently while the action branch reads the visual sequence, but future-video generation remains necessary~\cite{li2026lingbotva,zhang2026lingbotva2}. Fast-WAM then removes future-video generation at inference to reduce latency, retaining MoT's direct video K/V interface to action attention~\cite{yuan2026fastwam}.

However, these layer-wise MoT designs retain central limitations. First, each Video Expert layer directly exposes its dense visual features to the corresponding action layer by concatenating video and action keys/values, making task-relevant information and redundant details compete for attention. Prior work has found that action generation may allocate excessive attention to task-irrelevant visual regions, interfering with the use of task-relevant information and impairing task execution~\cite{li2026dtp}. Furthermore, reliance on task-irrelevant scene cues may weaken policy generalization under visual distribution shifts~\cite{sun2026afp}. Second, one-to-one layer coupling forces both experts to have equal depth, introducing redundant parameters and inference overhead. Excessive action-side capacity may favor memorizing specific scene patterns over learning generalizable environmental regularities, weakening generalization. WAMs therefore need a learnable interface that operates in a compact latent space and enables the Action Expert to allocate more attention to task-relevant visual features while allowing action supervision to shape visual representations. It should decouple expert depths, reducing redundant parameters and improving inference efficiency.

We introduce Bridge-of-Experts (BoE), a new architecture that overcomes these limitations, and instantiate it as \sysname{}, with an overview shown in Fig.~\ref{fig:bridgewam_teaser}. As illustrated in Fig.~\ref{fig:MoT_and_BoE}, BoE preserves modality-specific computation and connects the two experts through a compact set of Latent Bridge Queries (\queryname{}), rather than directly mixing video and action tokens. Within the Video Expert, 32 \queryname{} aggregate the visual, language, and proprioceptive priors needed for action generation into a shared, limited latent space, providing the Action Expert with latent features optimized for action generation through iterative learning. During end-to-end optimization, action-loss gradients flow through the \queryname{} into the Video Expert, guiding it to produce representations better suited to the Action Expert's learning and prediction. BoE decouples the depths of the two experts. \sysname{} combines a 30-layer Video Expert with a two-layer alternating cross--self-attention Action Expert: cross-attention retrieves information from the \queryname{}, while self-attention coordinates the action chunk. By retaining the backbone's representational capacity with a lightweight action head, this design balances efficiency and performance. Our contributions are as follows:

\begin{figure}[t]
      \centering
      \includegraphics[
      width=\linewidth,
      trim={0.5cm 3cm 0cm 2cm}
  ]{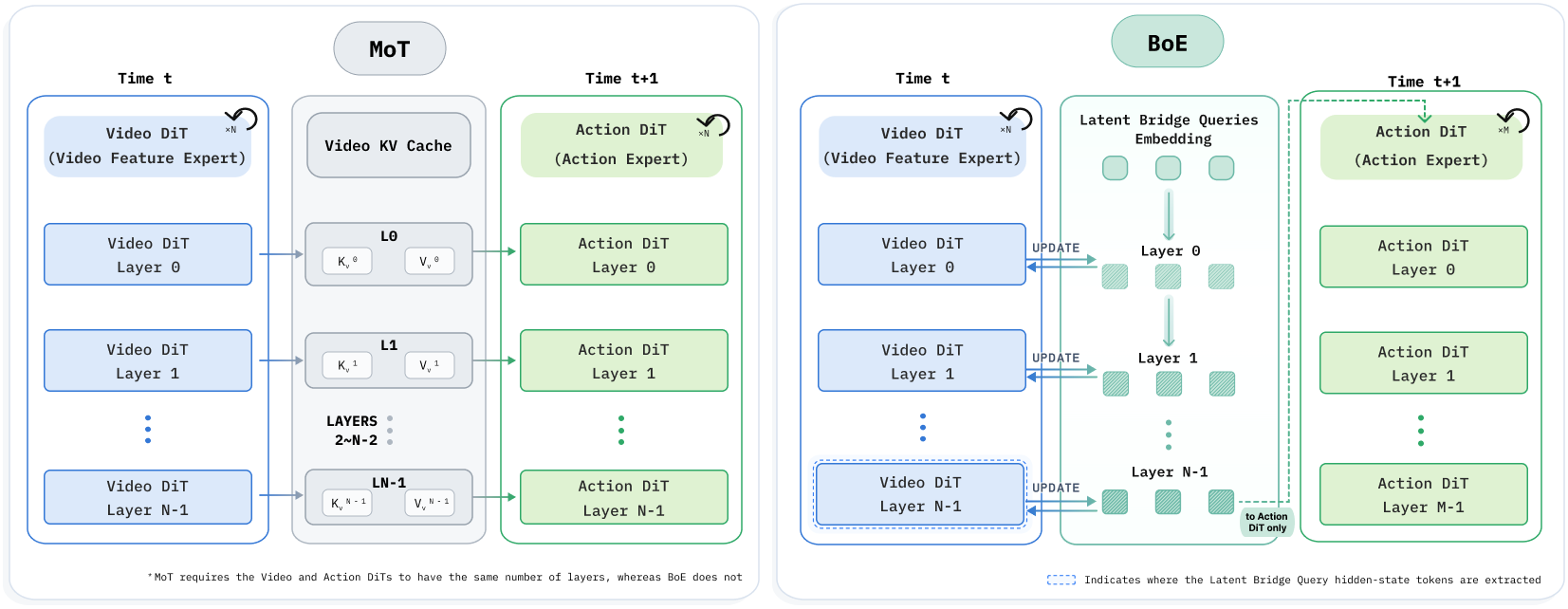}
    \caption{\textbf{Comparison of the BoE and MoT architectures.}
    BoE uses Latent Bridge Queries (LBQs) to transfer action-relevant visual features between experts.
    Unlike MoT's matched-depth coupling, BoE decouples expert depths, enabling \sysname{} to use only two layers in the Action DiT while retaining the full Video Expert for learning rich representations of visual information.}
      \label{fig:MoT_and_BoE}
  \end{figure}

\begin{itemize}
\item \textbf{We introduce BoE, a video--action interface built on a limited latent space.} BoE uses 32 \queryname{} to construct a limited latent space that aggregates visual, language, and proprioceptive priors into a unified latent representation tailored to the Action Expert. Joint video--action optimization sends action gradients through the \queryname{} to shape action-relevant video representations and trains the Action Expert to prioritize task-relevant visual information.

\item \textbf{We develop \sysname{} with decoupled expert depths to balance efficiency and performance.} \sysname{} pairs a 30-layer Video Expert with a two-layer alternating cross--self-attention Action Expert for LBQ readout and action-chunk coordination. The lightweight action head and reusable LBQ interface reduce redundancy in parameters and the cost of inference, yielding a $5.06$B-parameter \sysname{} that retains the video backbone's representational capacity while balancing performance and efficiency.

\item \textbf{We demonstrate that \sysname{} overcomes the above limitations of MoT} through simulation and real-world task evaluations and precise ablation studies. Meanwhile, we conduct systematic ablations, action-conditioned visual-routing visualizations, and efficiency analyses to examine the effects of the LBQ interface and Action Expert depth on task performance, the ability to generalize, and the cost of computation.
\end{itemize}

\section{Related Work}
  \paragraph{Generalist Vision--Language--Action Policies.}
  Generalist vision--language--action (VLA) policies map visual observations and language
  instructions directly to robot actions. RT-1 and RT-2 scale behavior cloning and vision--language
  pretraining for cross-task generalization~\citep{brohan2023rt1,zitkovich2023rt2}. OpenVLA,
  $\pi_0$, $\pi_{0.5}$, and GR00T N1 extend generalist control using large pretrained backbones with
  autoregressive, diffusion, or flow-matching action
  decoders~\citep{kim2024openvla,black2024pi0,pi05,bjorck2025gr00t}. Despite their strong ability to generalize semantically, these action-centric models learn
  physical dynamics only implicitly from demonstrations annotated with robot actions. This
  limitation motivates the development of World Action Models, which explicitly model how visual scenes evolve
  over time.

  \paragraph{World Action Models and Video--Action Coupling.}
  World Action Models (WAMs) jointly model visual dynamics and robot actions, providing dense
  supervision for scene evolution and physical
  interactions~\citep{gao2026dreamdojo,cen2025worldvla,bi2025motus,kim2026cosmospolicy}. DreamZero
  jointly denoises future-video and action latents, enabling cross-modal interaction throughout
  inference~\citep{ye2026dreamzero}. LingBot-VA and LingBot-VA 2.0 instead infer actions from
  predicted visual transitions through causal inverse
  dynamics~\citep{li2026lingbotva,zhang2026lingbotva2}. Fast-WAM reduces visual-imagination cost by
  training with future-video prediction but omitting future generation at inference, directly
  conditioning actions on latent Video DiT
  representations~\citep{yuan2026fastwam}. AHA-WAM further decouples their temporal roles through a
  low-frequency video planner and a high-frequency action executor~\citep{cai2026ahawam}. These WAMs primarily couple modalities through shared attention or by directly injecting
  dense video K/V representations.

  \paragraph{Cross-Modal Representation Transfer with Learnable Queries.}
  MetaQuery uses learnable queries to extract generation conditions from a frozen multimodal
  language model for transfer to a diffusion model~\citep{pan2025metaquery}. GenCeption inserts
  learnable tokens into a pretrained Video DiT to extract sparse structured predictions from its
  generative representations~\citep{wang2026genception}. These studies establish compact, learnable
  query interfaces for task-relevant information extraction, primarily through one-way condition
  transfer or task readout. Building on this insight, our Latent Bridge Queries (LBQs) use a shared
  set of learnable queries to construct a latent space for video--action associations.
  Unlike MetaQuery, which extracts generation conditions from a frozen multimodal language model,
  and GenCeption, which uses query tokens for sparse perception predictions, LBQs learn shared
  conditioning representations under joint video and action supervision, enabling the Video
  Expert's representation learning and the Action Expert's selective readout to adapt jointly.
  The Video Expert learns to write action-relevant physical representations into this shared
  latent space, while the Action Expert learns to selectively retrieve from the same space the
  information required for control and action generation under changing physical conditions.

\section{Methodology}

\subsection{Preliminaries}
\label{sec:preliminaries}

\paragraph{Embodied Policies.}
Following Motus~\cite{bi2025motus}, we describe embodied modeling
paradigms through their conditional distributions. Let $o_t$, $l$, and
$s_t$ denote the current visual observation, language instruction, and
robot proprioceptive state, respectively. We use $H$ and $K$ for the
action and video horizons, which cover the same prediction interval at
potentially different sampling rates:
\begin{center}
\renewcommand{\arraystretch}{1.08}
\begin{tabular}{@{}l@{\hspace{2em}}l@{}}
\toprule
\textbf{Paradigm} & \textbf{Conditional distribution} \\
\midrule
VLA         & $p(a_{t+1:t+H}\mid o_t,l)$ \\
WM          & $p(o_{t+1:t+K}\mid o_t,a_{t+1:t+H})$ \\
Joint WAM   & $p(o_{t+1:t+K},a_{t+1:t+H}\mid o_t,l,s_t)$ \\
IDM WAM     & $p(a_{t+1:t+H}\mid o_{t:t+K},l,s_t)$ \\
Fast-WAM    & $p(a_{t+1:t+H}\mid o_t,l,s_t)$ \\
\midrule
\sysname{} (ours) & $p(a_{t+1:t+H}\mid \mathbf{b}_t)$ \\
\bottomrule
\end{tabular}
\end{center}
We show canonical VLA and WM formulations and make state conditioning
explicit for the WAM policies considered here. VLA predicts actions;
WM predicts their visual consequences. Joint WAMs generate future
observations and actions together, whereas IDM WAMs condition actions
on visual transitions. Fast-WAM uses a VLA-like inference
interface while retaining video supervision during training~\cite{yuan2026fastwam}. The MoT-based WAM designs considered here couple their experts through
shared attention over video and action representations~\cite{bi2025motus,yuan2026fastwam}.
With one-to-one layer coupling, the Action Expert must match the Video
Expert's depth.

\paragraph{Bridge-of-Experts.}
We propose Bridge-of-Experts (BoE) to replace this coupling with a
compact latent interface. Within the Video Expert, Latent Bridge Queries
(\queryname{}) integrate the current observation, language instruction, and
robot state into a set of final hidden states:
\begin{equation}
    \mathbf{b}_t
    = f_V(o_t,l,s_t;\theta_V,\boldsymbol{\Phi})
    \in\mathbb{R}^{M\times d_V},
    \qquad M=32,
\label{eq:bridge_conditioning}
\end{equation}
where $o_t$, $l$, and $s_t$ denote the current visual observation, language
instruction, and robot state, respectively. $\theta_V$ denotes the Video
Expert parameters, and $\boldsymbol{\Phi}$ contains the learnable initial
LBQ embeddings. The encoding function $f_V$ maps these inputs and initial
queries to the final, input-conditioned LBQ representations $\mathbf{b}_t$.
Here, $M$ is the number of queries, and $d_V$ is the feature dimension of
each token in the Video Expert. The Action Expert receives all external
conditioning through $\mathbf{b}_t$, projected into its own token space.
This fixed-size interface aligns visual, semantic, and proprioceptive
information through joint video and action supervision without passing dense video tokens directly to the action branch during the conditioning and denoising processes.

During training, \queryname\ read current-frame features and text--state
context, while future-video queries read the \queryname\ to predict scene
evolution. The generation loss over all valid future frames also
updates the \queryname\ embeddings, encouraging predictive information that supports future-video generation to be injected into them. Action-loss gradients flow through the bridge into the Video
Expert, providing implicit feedback that encourages visual representations
useful for control. At inference, \queryname\ are computed once from the current
context and reused throughout action denoising. This design keeps action-conditioning information sources consistent across training and inference: LBQs
are constructed from current observations, language instructions, and proprioceptive
state, while future video provides training supervision without serving as an input
to the bridge. Future-video generation can therefore be omitted at inference without
changing how the Action Expert receives conditioning through LBQs. BoE therefore permits
independent expert depths while keeping the interface compact: \sysname{}
combines a 30-layer Video Expert with a two-layer alternating
cross--self-attention Action Expert for compact action conditioning every
denoising step.

\begin{figure}[t]
      \centering
      \includegraphics[
      width=\linewidth,
      trim={0.1cm 1cm 0cm 2cm},
      clip
  ]{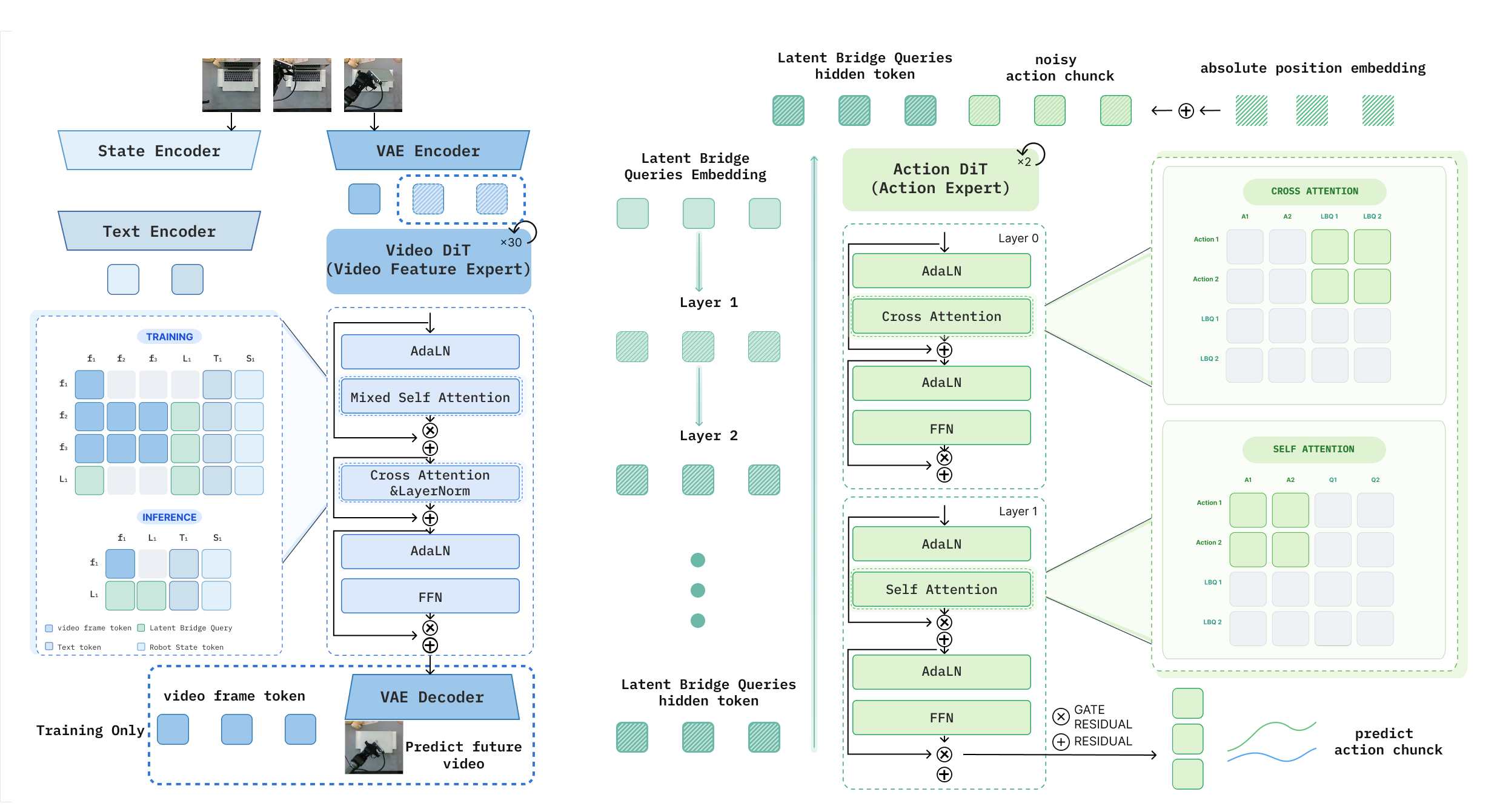}
      \caption{\textbf{Overview of the BridgeWAM architecture.}
      Latent Bridge Queries (\queryname{}) first aggregate action-relevant  representations from the Video DiT and are read by the
      Action DiT to guide action prediction, forming a compact bridge through which information flows from video to action.}
      \label{fig:bridgewam_architecture}
  \end{figure}

\subsection{Learning and Reading the Latent Bridge}
\label{sec:video_lbq}
\label{sec:action_lbq}
\label{sec:training_objective}

In Fig.~\ref{fig:bridgewam_architecture}, we initialize LBQs with
$\mathbf{B}^{0}=\boldsymbol{\Phi}$ and propagate them through the Video DiT.
At layer $\ell$, $\mathbf{V}^{\ell}=[\mathbf{V}_0^{\ell};\mathbf{V}_f^{\ell}]$
contains current-frame and future-video tokens, and $\mathbf{B}^{\ell}$
denotes LBQ states. Future-video queries read video and LBQ keys/values,
whereas LBQ queries read only current-frame and LBQ keys/values, excluding
future-video keys/values from LBQ updates at every Video DiT layer:
\begin{equation}
\begin{aligned}
    \mathbf{R}_{V_f}^{\ell}
    &=\operatorname{Attn}\!\left(
        \mathbf{Q}_{V_f}^{\ell},
        [\mathbf{K}_{V}^{\ell};\mathbf{K}_B^{\ell}],
        [\mathbf{U}_{V}^{\ell};\mathbf{U}_B^{\ell}];\mathbf{M}_f
      \right), \\
    \mathbf{R}_B^{\ell}
    &=\operatorname{Attn}\!\left(
        \mathbf{Q}_B^{\ell},
        [\mathbf{K}_{V_0}^{\ell};\mathbf{K}_B^{\ell}],
        [\mathbf{U}_{V_0}^{\ell};\mathbf{U}_B^{\ell}]
      \right).
\end{aligned}
\label{eq:video_bridge_update}
\end{equation}
Here, $\mathbf{Q},\mathbf{K},\mathbf{U}$ denote queries, keys, and values computed with shared Video DiT projections; subscripts identify source tokens, and $\cdot;\cdot$ denotes concatenation. $\mathbf{M}_f$ preserves video masking while allowing future queries to read all LBQs, whose states are updated through residual readouts, text--state cross-attention, and an FFN. Video tokens use 3D RoPE and LBQs use identity RoPE; the K/V restriction blocks LBQ access to future-video tokens. The final bridge $\mathbf{b}_t=\mathbf{B}^{L_V}$ is projected into the Action DiT space as $\widehat{\mathbf{B}}=\Pi(\mathbf{b}_t)$; action queries read these LBQ keys/values without directly receiving text--state tokens or dense video K/V. We jointly fine-tune both experts and the learnable LBQ embeddings $\boldsymbol{\Phi}$ with equally weighted video and action flow-matching losses. Video loss updates the Video Expert and $\boldsymbol{\Phi}$, while action loss updates the Action Expert, $\Pi$, $\boldsymbol{\Phi}$, and the Video Expert through the differentiable LBQ pathway. Thus, each LBQ update combines contributions from both losses, enabling action supervision to guide the Video Expert toward visual representations that support task-relevant perception, action prediction, and closed-loop control.

\subsection{Alternating Cross--Self Action Head}
\label{sec:alternating_action_head}

Following the alternating design in GR00T N1~\cite{bjorck2025gr00t},
we use a two-layer Action DiT: cross-attention retrieves information from
LBQs, then self-attention coordinates actions within the chunk.
For noisy actions $\{\mathbf{a}_{\sigma,i}\}_{i=0}^{H-1}$, we form
$\mathbf{x}=[\mathbf{x}_0;\ldots;\mathbf{x}_{H-1}]$ with learned absolute
positions that distinguish temporal slots within the action chunk before
cross-attention is applied:
\begin{equation}
    \mathbf{x}_i=\operatorname{Linear}_{\mathrm{in}}(\mathbf{a}_{\sigma,i})
                 +\mathbf{p}_i,
    \qquad \mathbf{p}_i=\operatorname{Emb}_{\mathrm{pos}}(i).
\label{eq:action_position_embedding}
\end{equation}
Each action receives a learned embedding of its position within the chunk
to distinguish temporal slots before cross-attention, while 1D RoPE in
self-attention encodes relative positions between actions.
The full mapping from action tokens to predicted action flow is
\begin{equation}
\begin{aligned}
    \mathbf{z}
    &=\operatorname{SelfBlock}_{t_A}\!\left(
        \operatorname{CrossBlock}_{t_A}(\mathbf{x},\widehat{\mathbf{B}})
      \right), \\
    \widehat{\mathbf{u}}_A&=\operatorname{Linear}_{\mathrm{out}}(\mathbf{z}).
\end{aligned}
\label{eq:alternating_action_head}
\end{equation}
Each block comprises attention followed by an FFN, with residual connections
and AdaLN conditioned on the diffusion timestep $t_A$. Self-attention and
FFN residual branches use timestep-conditioned gates; cross-attention uses
an ungated residual. The cross block uses action queries and LBQ keys/values;
the self block uses only action tokens. The output $\widehat{\mathbf{u}}_A$
predicts action flow, with the same LBQs reused as the action-conditioning
context at every denoising step.

  \section{Experiments}
  \label{sec:experiments}

\providecommand{\sysname}{\textsc{LQ-WAM}}
\providecommand{\cmark}{\ding{51}}
\providecommand{\xmark}{\ding{55}}

\begin{table}[t]
    \centering
    \small
    \setlength{\abovecaptionskip}{0pt}
    \caption{Results on LIBERO and LIBERO-Pro. We report task success rates (\%) on the Spatial, Object, Goal, and Long-Horizon suites, together with the average success rate across these suites. }
    \label{tab:libero_main}
    \resizebox{\linewidth}{!}{
        \begin{tabular}{l|c|ccccc|ccccc}
            \toprule
             \textbf{Method}
                &  \textbf{Embodied PT.}
                & \multicolumn{5}{c|}{\textbf{LIBERO}}
                & \multicolumn{5}{c}{\textbf{LIBERO-Pro (zero-shot)}} \\
            \cmidrule(lr){3-7}\cmidrule(lr){8-12}
                &
                & \textbf{Spatial} $\uparrow$
                & \textbf{Object} $\uparrow$
                & \textbf{Goal} $\uparrow$
                & \textbf{Long} $\uparrow$
                & \textbf{Avg.} $\uparrow$
                & \textbf{Spatial} $\uparrow$
                & \textbf{Object} $\uparrow$
                & \textbf{Goal} $\uparrow$
                & \textbf{Long} $\uparrow$
                & \textbf{Avg.} $\uparrow$ \\
            \midrule

            \textbf{OpenVLA} \cite{kim2024openvla}
                & \cmark
                & 84.7 & 88.4 & 79.2 & 53.7 & 76.5
                & 40.5 & 32.0 & 25.0 & 20.5 & 29.5 \\

            \textbf{$\pi_{0}$} \cite{black2024pi0}
                & \cmark
                & 96.8 & 98.8 & 95.8 & 85.2 & 94.1
                & 52.0 & 33.2 & 54.6 & 33.2 & 43.2\textsuperscript{*} \\

            \textbf{$\pi_{0.5}$} \cite{pi05}
                & \cmark
                & 98.8 & 98.2 & 98.0 & 92.4 & 96.9
                & 55.6 & 52.2 &  48.0 &      \textbf{57.0} & 53.0\textsuperscript{*} \\

            \textbf{LingBot-VA} \cite{li2026lingbotva}
                & \cmark
                & 98.5 & 99.6 & 97.2 & \textbf{98.5} & 98.5
                & -- & -- & -- & -- & --  \\

            \textbf{Motus} \cite{bi2025motus}
                & \cmark
                & 96.8 & 99.8 & 96.6 & 97.6 & 97.7
                & -- & -- & -- & -- & --  \\

            \midrule
            \multicolumn{12}{l}{\textit{World Action Model}} \\

            \textbf{Fast-WAM} \cite{yuan2026fastwam}
                & \xmark
                & 97.8 & 99.4 & 95.4 & 94.6 & 96.8
                & 57.6 & 49.2 & 58.1 & 40.8 & 51.4 \\

            \textbf{Faster-WAM} \cite{zhao2026faster}
                & \xmark
                & \textbf{99.6} & 99.8 & \textbf{98.2} & 98.2 & \textbf{99.0}
                & \textbf{59.6} & 52.8 &\textbf{59.6} & 45.8 & \textbf{54.5} \\

            \rowcolor{bridgewamrow!60!white}
            \textbf{\sysname{} (Ours)}
                & \xmark
                & 99.0 & \textbf{100.0} & 97.4 & 96.6 & 98.3
                & 57.8 & \textbf{55.1} & 58.4 & 43.5 & 53.7 \\

            \bottomrule
        \end{tabular}

}
  \par\vspace{2pt}
  \noindent
  \parbox{\linewidth}{%
      \raggedright
      \scriptsize
      \textsuperscript{*} Reproduced with hyperparameter settings identical to those in the official LIBERO-Pro repository.%
  }
  \par
    \vspace{0.5em}

    \caption{Zero-shot results on LIBERO-Plus. We report task success rates (\%) under seven types of perturbations and their average. ``Embodied PT.'' indicates whether embodied pre-training is used.}
    \label{tab:libero_plus}
    \resizebox{\linewidth}{!}{
        \begin{tabular}{l|c|cccccccc}
            \toprule
            \textbf{Method}
                & \textbf{Embodied PT.}
                & \multicolumn{8}{c}{\textbf{LIBERO-Plus (zero-shot)}} \\
            \cmidrule(lr){3-10}
                &
                & \textbf{Camera} $\uparrow$
                & \textbf{Robot} $\uparrow$
                & \textbf{Lang.} $\uparrow$
                & \textbf{Light} $\uparrow$
                & \textbf{Backg.} $\uparrow$
                & \textbf{Noise} $\uparrow$
                & \textbf{Layout} $\uparrow$
                & \textbf{Avg.} $\uparrow$ \\
            \midrule

            \multicolumn{10}{l}{\textit{Vision Language Model}} \\

            \textbf{OpenVLA} \cite{kim2024openvla}
                & \cmark
                & 0.6 & 3.9 & 38.7 & 14.2 & 38.4 & 5.4 & 42.5 & 19.7 \\
            
            \textbf{OpenVLA-OFT}  \cite{kim2025finetuning}
                & \cmark
                & 56.4 & 31.9 & 79.5 & 88.7 & 93.3 & 75.8 & 74.2 & 69.6 \\

            \textbf{$\pi_{0}$} \cite{black2024pi0}
                & \cmark
                & 13.8 & 6.0 & 58.8 & 85.0 & 81.4 & \textbf{79.0} & 68.9 & 53.6 \\

            \midrule
            \multicolumn{10}{l}{\textit{World Action Model}} \\

            \textbf{Fast-WAM} \cite{yuan2026fastwam}
                & \xmark
                & 15.4 & 44.4 & 68.5 & 79.0 & 52.6 & 37.3 & 61.5 & 49.5 \\

            \textbf{OpenWAM-$\alpha$}  \cite{wang2026openwam}
                & \xmark
                & 33.8 &\textbf{76.1} &88.0 &\textbf{97.0} &\textbf{87.1} &39.8 &77.5 &69.2 \\

            \textbf{Faster-WAM} \cite{zhao2026faster}
                & \xmark
                & 53.8 & 71.6 & \textbf{94.7} & 96.3 & 61.3 & 63.6 & \textbf{79.1} & \textbf{73.6}\\
            \rowcolor{bridgewamrow!60!white}
            \textbf{\sysname{} (Ours)}
                & \xmark
                & \textbf{56.5} & 60.5 & 85.1 & 92.2 & 69.2 & 73.6 & 75.7 & 72.2 \\

            \bottomrule
        \end{tabular}
        }
\end{table}

  This section systematically evaluates \sysname{} in terms of effective visual representation use and decoupled expert
  depths, examining the properties of BoE described in the Introduction. First, quantitative experiments in simulation
  and on real robots (Secs.~\ref{sec:simulation_results} and~\ref{sec:real_world_results}) test whether an LBQ interface
  built on a limited latent space and a lightweight Action Expert can maintain effective task execution and generalize
  under environmental perturbations and distribution shifts. Meanwhile, ablation studies
  (Sec.~\ref{sec:ablation_results}) examine the joint adaptation of visual representations and action prediction, the
  aggregation of multimodal priors in a limited latent space, and the effects of the Action Expert's readout mechanism
  on task performance. Additionally, efficiency experiments (Sec.~\ref{sec:ablation_results}) compare parameter counts,
  computational cost, and inference latency across simulation-input profiling and real-robot deployment settings,
  testing whether decoupled expert depths and LBQ conditioning reuse translate into practical computational savings and,
  together with the quantitative results, assessing the balance between performance, generalization, and inference
  efficiency. Finally, qualitative analysis (Sec.~\ref{sec:qualitative_results}) illustrates how BoE guides the Action
  Expert to assign greater attention to visual features of target objects and task destinations during action
  prediction, and examines the stability of this guidance under different perturbations.

  \subsection{Experimental Setup}
  \label{sec:experimental_setup}

  We evaluate \sysname\  on LIBERO\cite{liu2023libero}, LIBERO-Plus\cite{fei25libero-plus}, LIBERO-Pro\cite{zhou2025liberopro}, RoboTwin 2.0\cite{tianxing2025robotwin}, and real-world manipulation tasks. Unless otherwise specified, \sysname\  initializes its 30-layer Video Expert from Wan2.2-
  TI2V-5B~\cite{wan2025} and employs 32 Latent Bridge Queries together with a lightweight two-layer alternating cross--self-attention Action Expert. The video and action objectives are jointly optimized with
  equal weights. The model is trained for 10 epochs on LIBERO and 5 epochs on RoboTwin 2.0. LIBERO experiments use two RGB camera views, whereas RoboTwin 2.0 uses three views. During evaluation, we use 10 action-denoising steps and report task success rates in accordance with the evaluation protocol and corresponding task definitions of each benchmark.

  {\setlength{\textfloatsep}{2pt}
   \setlength{\intextsep}{2pt}
   \setlength{\abovecaptionskip}{3pt}
   \providecommand{\sysname}{\textsc{LQ-WAM}}
\providecommand{\cmark}{\ding{51}}
\providecommand{\xmark}{\ding{55}}

\begin{table}[t]
\centering
\caption{Fine-grained task comparison on RoboTwin 2.0 Simulation
(Clean vs. Randomized).}
\label{tab:robotwin_main}
\scriptsize
\resizebox{\textwidth}{!}{
\begin{tabular}{l|cc|cc|cc|cc|cc|cc|cc|>{\columncolor{bridgewamrow!60!white}}c>{\columncolor{bridgewamrow!60!white}}c}
\toprule
\textbf{Task}
& \multicolumn{2}{c|}{$\pi_{0}$}
& \multicolumn{2}{c|}{$\pi_{0.5}$}
& \multicolumn{2}{c|}{Motus}
& \multicolumn{2}{c|}{Motus from WAN2.2}
& \multicolumn{2}{c|}{LingBot-VA}
& \multicolumn{2}{c|}{Fast-WAM}
& \multicolumn{2}{c|}{Faster-WAM}
& \multicolumn{2}{>{\columncolor{bridgewamrow!60!white}}c}{BridgeWAM (Ours)} \\
& \multicolumn{2}{c|}{\cite{black2024pi0}}
& \multicolumn{2}{c|}{\cite{pi05}}
& \multicolumn{2}{c|}{\cite{bi2025motus}}
& \multicolumn{2}{c|}{\cite{bi2025motus}}
& \multicolumn{2}{c|}{\cite{li2026lingbotva}}
& \multicolumn{2}{c|}{\cite{yuan2026fastwam}}
& \multicolumn{2}{c|}{\cite{zhao2026faster}}
& \multicolumn{2}{>{\columncolor{bridgewamrow!60!white}}c}{} \\
& C & R & C & R & C & R & C & R & C & R
& C & R & C & R & C & R \\
\midrule

Adjust Bottle
& 99 & 95 & \textbf{100} & 99 & 89 & 93 & 99 & 97 & 90 & 94 
& \textbf{100} & \textbf{100} & \textbf{100} & 99 & \textbf{100} & 99 \\

Blocks Ranking RGB
& 80 & 63 & 92 & 85 & 99 & 97 & 92 & 88 & 99 & 98 
& \textbf{100} & \textbf{100} & \textbf{100} & \textbf{100} & 99 & 99 \\

Click Bell
& 71 & 48 & 99 & 66 & \textbf{100} & \textbf{100} & \textbf{100} & \textbf{100} & \textbf{100} & \textbf{100} 
& \textbf{100} & \textbf{100} & \textbf{100} & \textbf{100} & 99 & 99 \\

Grab Roller
& 98 & 94 & \textbf{100} & \textbf{100} & \textbf{100} & \textbf{100} & \textbf{100} & \textbf{100} & \textbf{100} & \textbf{100} 
& \textbf{100} & \textbf{100} & \textbf{100} & \textbf{100} & \textbf{100} & \textbf{100} \\

Handover Mic
& 97 & 97 & 98 & 97 & 78 & 63 & 98 & 95 & 94 & 96 
& 99 & \textbf{100} & \textbf{100} & 99 & \textbf{100} & \textbf{100} \\

Lift Pot
& 80 & 72 & 96 & 85 & 96 & 99 & 90 & 87 & \textbf{100} & 99 
& \textbf{100} & \textbf{100} & \textbf{100} & \textbf{100} & \textbf{100} & \textbf{100} \\

Shake Bottle Horizontally
& 98 & 92 & 99 & 99 & \textbf{100} & 98 & \textbf{100} & 97 & \textbf{100} & 99 
& \textbf{100} & \textbf{100} & \textbf{100} & \textbf{100} & \textbf{100} & \textbf{100} \\

Shake Bottle
& 94 & 91 & 99 & 97 & \textbf{100} & 97 & \textbf{100} & 96 & \textbf{100} & 97 
& \textbf{100} & \textbf{100} & \textbf{100} & \textbf{100} & \textbf{100} & \textbf{100} \\

Move Pillbottle Pad
& 67 & 46 & 84 & 61 & 93 & 96 & 83 & 83 & 99 & 99 
& \textbf{100} & 99 & 98 & 99 & 99 & \textbf{100} \\

Move Playingcard Away
& 74 & 65 & 96 & 84 & \textbf{100} & 96 & 50 & 47 & \textbf{100} & 99 
& \textbf{100} & \textbf{100} & \textbf{100} & \textbf{100} & 98 & \textbf{100} \\

......(50 tasks)  & & & & & & & & & & & & & & & & \\

Click Alarmclock
& 77 & 68 & 98 & 89 & \textbf{100} & \textbf{100} & \textbf{100} & 99 & 99 & \textbf{100} 
& \textbf{100} & \textbf{100} & 99 & \textbf{100} & 97 & 99 \\

Dump Bin Bigbin
& 88 & 83 & 92 & 97 & 95 & 91 & 94 & 96 & 89 & 96 
& 97 & 96 & 97 & \textbf{98} & \textbf{98} & 97 \\

Open Laptop
& 71 & 81 & 90 & 96 & 95 & 91 & 89 & 89 & 92 & 94 
& 98 & \textbf{100} & 95 & 99 & \textbf{99} & \textbf{100} \\

Place Container Plate
& 97 & 92 & \textbf{99} & 95 & 98 & 99 & 97 & \textbf{100} & \textbf{99} & 97 
& 96 & \textbf{100} & 98 & 97 & \textbf{99} & \textbf{100} \\

Place Burger Fries
& 81 & 76 & 94 & 87 & \textbf{98} & 98 & 95 & 90 & 97 & 95 
& 96 & \textbf{99} & \textbf{98} & 97 & \textbf{98} & 98 \\

Place Object Stand
& 82 & 68 & 91 & 85 & 98 & 97 & 91 & 93 & \textbf{99} & 96 
& 90 & 94 & 95 & 96 & 96 & \textbf{98} \\

Place Phone Stand
& 49 & 53 & 81 & 81 & 87 & 86 & 80 & 78 & 97 & 97 
& 97 & \textbf{99} & \textbf{100} & 96 & 98 & 95 \\

Place Shoe
& 76 & 76 & 92 & 93 & \textbf{99} & 97 & 95 & 92 & 98 & 98 
& 96 & \textbf{99} & 97 & 98 & 97 & 96 \\

Stack Bowls Two
& 94 & 95 & 95 & 96 & 98 & \textbf{98} & 98 & \textbf{98} & 94 & \textbf{98} 
& 92 & \textbf{98} & 98 & 96 & \textbf{99} & 97 \\

Stack Blocks Two
& 93 & 79 & 97 & \textbf{100} & \textbf{100} & 98 & 96 & 94 & \textbf{100} & 98 
& \textbf{100} & \textbf{100} & \textbf{100} & \textbf{100} & \textbf{100} & 99 \\

Stack Bowls Three
& 77 & 75 & 77 & 71 & 79 & 87 & \textbf{90} & 74 & 86 & 83 
& 80 & 81 & 83 & 78 & \textbf{88} & \textbf{90} \\

\midrule

\textbf{Average (\%)}
& 65.92 & 58.40
& 82.74 & 76.76
& 88.66 & 87.02
& 77.56 & 77.00
& \textbf{92.90} & 91.50
& 91.88 & 91.78
& 92.78 & \textbf{92.26}
& 89.84 & 90.22 \\

\bottomrule
\end{tabular}
}
\end{table}

   \begin{figure}[t]
       \centering
       \includegraphics[width=\linewidth,trim={0cm 2cm 0cm 1.5cm}]{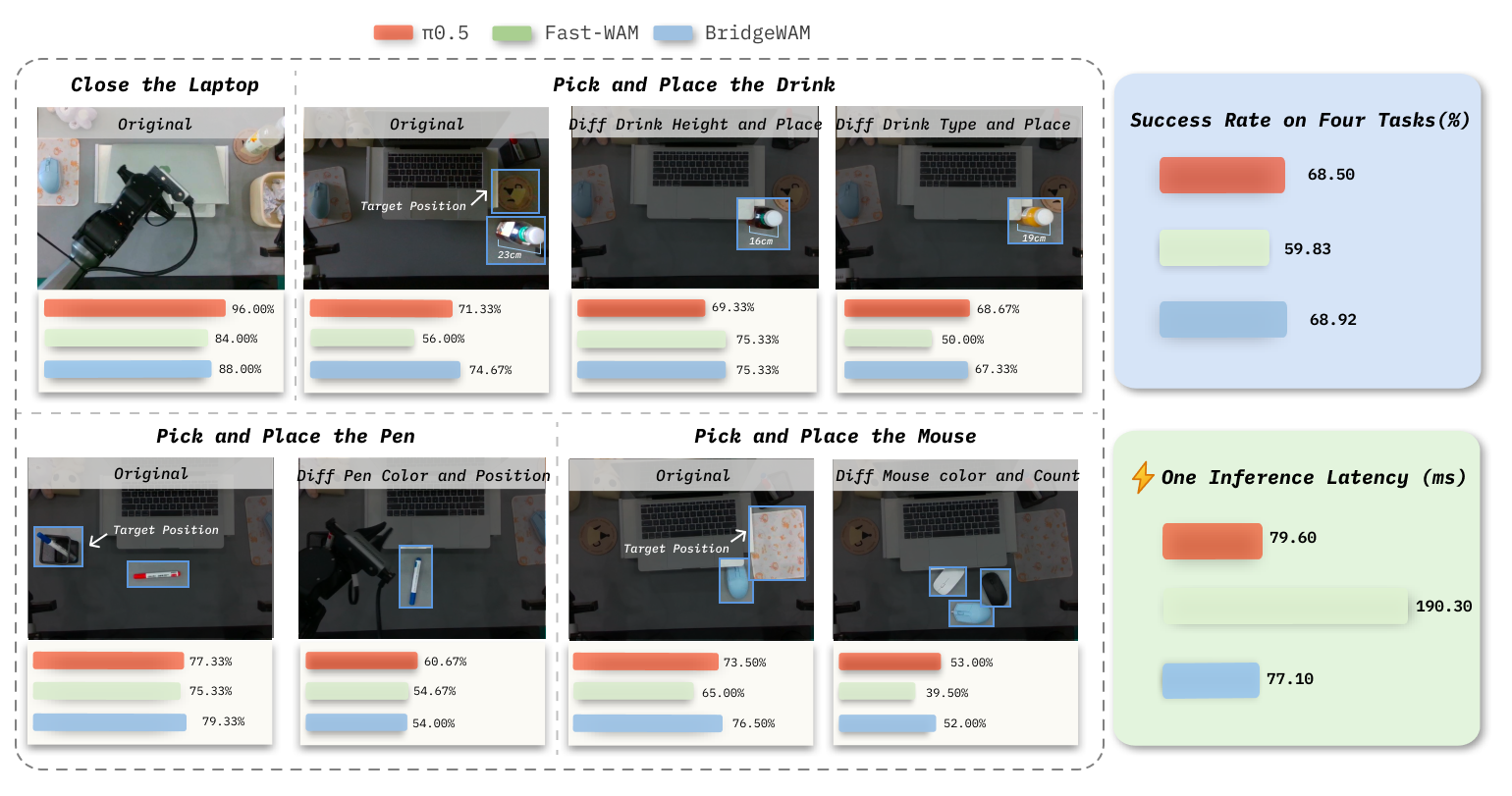}
       \caption{\textbf{Real-world manipulation tasks and results.}
       We compare $\pi_{0.5}$, Fast-WAM, and \sysname{} on closing a laptop
       and picking and placing a drink, a pen, and a mouse.
       The task panels show the original settings and variations in object
       height, type, color, position, or count, together with success rates (\%).
       The right panels summarize success rates and inference latency (ms).}
       \label{fig:real_world_results}
   \end{figure}}
  \subsection{Quantitative Evaluation in Simulation}
  \label{sec:simulation_results}
  \label{sec:libero_results}

  \paragraph{LIBERO Benchmarks.}

  \sysname{} averages $98.25\%$ on LIBERO ($1.45$ points above Fast-WAM), $100\%$ on LIBERO-Object, and $53.7\%$ on LIBERO-Pro (Table~\ref{tab:libero_main}). On LIBERO-Plus, it raises Fast-WAM's average from $49.5\%$ to $72.2\%$, with the best Camera score ($56.5\%$; Table~\ref{tab:libero_plus}).

  \paragraph{RoboTwin 2.0.}
  \label{sec:robotwin_results}

  We train for five epochs (29,355 steps) on Fast-WAM's 27,500 demonstrations (2,500 clean; 25,000 randomized), evaluating 50 tasks with 100 trials per task and setting (Table~\ref{tab:robotwin_main}). Under randomization, \sysname{} scores $90\%$ on \textit{Stack Bowls Three} (Faster-WAM: $78\%$; LingBot-VA: $83\%$), $98\%$ on \textit{Place Object Stand}, and $100\%$ on \textit{Move Pillbottle Pad}. The success rate on \textit{Handover Mic}, \textit{Lift Pot}, and \textit{Open Laptop} is near-perfect under both of the evaluated settings.

  \paragraph{Analysis.}
  Across LIBERO, LIBERO-Plus, LIBERO-Pro, and RoboTwin 2.0, \sysname{} supports effective task execution and strong generalization with an ultralight two-layer Action Expert. Together with the efficiency analysis in Sec.~\ref{sec:ablation_results}, these results highlight BoE's performance--efficiency balance: \sysname{} uses $5.06$B parameters versus Faster-WAM's $5.56$B, approximately $9.0\%$ fewer (Fig.~\ref{fig:bridgewam_teaser}), and requires $54.67\%$ fewer profiled FLOPs under the respective native configurations (Table~\ref{tab:inference_efficiency}). The recorded H20Z latency is $88.05$\,ms versus $321.16$\,ms, approximately $1/3.65$ of Faster-WAM's. Within this performance--efficiency trade-off, Faster-WAM achieves higher aggregate success rates in these comparisons, potentially benefiting from future-aware conditioning beyond current-observation features~\citep{zhao2026faster}, although the present comparison does not isolate this effect. These results demonstrate that BoE balances effective task execution with reductions in both computational cost and inference latency.\looseness=-1

  {\setlength{\intextsep}{4pt}
  
\providecommand{\sysname}{\textsc{BridgeWAM}}

\begin{table}[t]
\centering
\setlength{\tabcolsep}{3pt}

\caption{LIBERO and LIBERO-Plus ablations of the interface and video--action coupling. Task success rates (\%) are reported.}
\label{tab:interface_ablation}
\resizebox{\linewidth}{!}{%
\begin{tabular}{l|ccccc|ccccc}
\toprule
\textbf{Method}
& \multicolumn{5}{c|}{\textbf{LIBERO}}
& \multicolumn{5}{c}{\textbf{LIBERO-Plus (Zero-shot)}} \\
\Xcline{2-6}{0.4pt}
\Xcline{7-11}{0.4pt}
& \textbf{Spatial} $\uparrow$
& \textbf{Object} $\uparrow$
& \textbf{Goal} $\uparrow$
& \textbf{Long} $\uparrow$
& \textbf{Avg.} $\uparrow$
& \textbf{Spatial} $\uparrow$
& \textbf{Object} $\uparrow$
& \textbf{Goal} $\uparrow$
& \textbf{Long} $\uparrow$
& \textbf{Avg.} $\uparrow$ \\
\midrule
\textbf{Fast-WAM}
&  97.8 &  99.4 &  95.4 &  94.6 &  96.80
&  52.21 &  69.06 &  37.32 &  40.06 &  49.54 \\
\rowcolor{bridgewamrow!60!white}
\textbf{\sysname{} (Ours)}
&  \textbf{99.0} &  \textbf{100.0} &  97.4 &  96.6 &  \textbf{98.25}
&  \textbf{77.64} &  \textbf{84.47} &  \textbf{60.29} &  \textbf{67.17} &  \textbf{72.24} \\
\midrule
\textbf{Frozen-Video-DiT}
&  87.8 &  97.0 &  84.8 &  72.4 &  85.50
&  32.22 &  38.52 &  15.44 &  16.36 &  25.48 \\
\textbf{MoT-readout}
&  98.0 &  99.4 &  97.4 &  96.4 &  97.80
&  71.57 &  80.70 &  36.09 &  54.78 &  60.48 \\
\textbf{Text-state-LBQ-input}
&  96.4 &  99.2 &  97.0 &  96.2 &  97.20
&  64.70 &  79.98 &  44.11 &  49.07 &  59.29 \\
\midrule
\textbf{\sysname\ Joint}
&  97.4 &  99.6 &  96.6 &  \textbf{96.8} &  97.60
&  76.52 &  72.12 &  46.51 &  59.59 &  63.41 \\
\textbf{\sysname\ IDM}
&  98.8 &  97.8 &  \textbf{99.0} &  96.6 &  98.05
&  76.23 &  71.49 &  58.55 &  64.19 &  67.45 \\
\bottomrule
\end{tabular}
}
\end{table}

\begin{table}[t]
\centering
\setlength{\tabcolsep}{3pt}

\caption{Ablations of action-head depth and LBQ sequence length on LIBERO and LIBERO-Plus.}
\label{tab:action_head_depth}
\label{tab:lbq_length_ablation}
\resizebox{\linewidth}{!}{%
\begin{tabular}{l|ccccc|ccccc}
\toprule
\textbf{Method}
& \multicolumn{5}{c|}{\textbf{LIBERO}}
& \multicolumn{5}{c}{\textbf{LIBERO-Plus (Zero-shot)}} \\
\Xcline{2-6}{0.4pt}
\Xcline{7-11}{0.4pt}
& \textbf{Spatial} $\uparrow$
& \textbf{Object} $\uparrow$
& \textbf{Goal} $\uparrow$
& \textbf{Long} $\uparrow$
& \textbf{Avg.} $\uparrow$
& \textbf{Spatial} $\uparrow$
& \textbf{Object} $\uparrow$
& \textbf{Goal} $\uparrow$
& \textbf{Long} $\uparrow$
& \textbf{Avg.} $\uparrow$ \\
\midrule
\textbf{Fast-WAM}
&  97.8 &  99.4 &  95.4 &  94.6 &  96.80
&  52.21 &  69.06 &  37.32 &  40.06 &  49.54 \\
\rowcolor{bridgewamrow!60!white}
\textbf{\sysname{} (Ours)}
&  \textbf{99.0} &  \textbf{100.0} &  97.4 &  96.6 &  \textbf{98.25}
&  77.64 &  84.47 &  60.29 &  67.17 &  72.24 \\
\midrule
\textbf{4 layers}
&  97.2 & \textbf{100.0} &  97.2 &  96.6 &  97.75
&  67.44 &  84.63 &  54.30 &  60.26 &  66.56 \\
\textbf{8 layers}
&  97.4 &  99.8 &  97.6 &  \textbf{96.8} &  97.90
&  70.69 &  83.76 &  49.83 &  58.44 &  65.50 \\
\textbf{16 layers}
&  98.8 &  99.4 &  96.2 &  93.4 &  96.95
&  67.94 &  80.70 &  52.37 &  62.13 &  65.66 \\
\textbf{30 layers (self-cross)}
&  97.8 & \textbf{100.0}&  97.6  &  94.8 &  97.55
&  69.40 &  78.04 &  60.09 &  64.31 &  67.89 \\
\midrule
\textbf{1 LBQ}
& 98.60 & 99.60 & 95.40 & 92.00 & 96.40
& 71.86 & 82.53 & 50.17 & 57.64 & 65.36 \\
\textbf{8 LBQs}
& 98.00 & 99.80 & 97.20 & 96.20 & 97.80
& 72.19 & 79.75 & 53.07 & 59.90 & 66.06 \\
\textbf{16 LBQs}
& 98.20 & 99.40 & \textbf{98.40} & 96.40 & 98.10
& 75.60 & 81.25 & 52.72 & 59.71 & 67.12 \\
\textbf{24 LBQs}
& 98.20 & 99.60 & 96.60 & 95.60 & 97.50
& \textbf{81.64} & 82.92 & 59.36 & 67.41 & \textbf{72.63} \\
\textbf{36 LBQs}
& 97.60 & 99.80 & 97.40 & 96.60 & 97.85
& 74.10 & \textbf{85.39} & \textbf{61.17} & \textbf{67.96} & 72.05 \\
\textbf{64 LBQs}
& 98.40 & 99.80 & 98.20 & 94.80 & 97.80
& 65.11 & 76.05 & 49.44 & 61.81 & 62.98 \\
\bottomrule
\end{tabular}%
}
\end{table}

\begin{table}[t]
    \centering
    \setlength{\tabcolsep}{3pt}

    \caption{Inference efficiency on LIBERO with 30-layer Video DiTs, H20Z GPUs, \texttt{bfloat16}, an action horizon of 32, and 10 denoising steps. Appendix~\ref{sec:appendix_efficiency} provides details of the profiling scope.}
    \label{tab:inference_efficiency}
    \resizebox{\linewidth}{!}{
        \begin{tabular}{lcccccccc}
            \toprule
            \textbf{Method}
                & \multicolumn{2}{c}{\textbf{Layers}}
                & \multicolumn{3}{c}{\textbf{FLOPs (T)}}
                & \multicolumn{2}{c}{\textbf{Acceleration}}
                & \makecell{\textbf{Peak Mem.}\\\textbf{(MiB)}} \\
            \cmidrule(lr){2-3}\cmidrule(lr){4-6}\cmidrule(lr){7-8}
                & \textbf{Video}
                & \textbf{Action}
                & \textbf{Video} $\downarrow$
                & \textbf{Action} $\downarrow$
                & \textbf{Total} $\downarrow$
                & \textbf{Latency (ms)} $\downarrow$
                & \textbf{Speedup} $\uparrow$
                & \\
            \midrule
            \textbf{Fast-WAM} \cite{yuan2026fastwam}
                & 30 & 30
                & \textbf{1.005} & 1.043 & 2.048
                & 405.91 & $1.00\times$
                & 23,773.86 \\
            \textbf{Faster-WAM} \cite{zhao2026faster}
                & 30 & 30
                & 2.745 & 0.476 & 3.221
                & 321.16 & 1.26$\times$
                & 22,962.00 \\
            \rowcolor{bridgewamrow!60!white}
            \textbf{\sysname{}}
                & 30 & 2
                & 1.432 & \textbf{0.028} & \textbf{1.460}
                & \textbf{88.05} & \textbf{4.61}$\times$
                & \textbf{21,912.21} \\
            \bottomrule
        \end{tabular}
    }
\end{table}

\begin{table}[t]
\centering
\small
\setlength{\tabcolsep}{5pt}
\caption{Zero-shot IDM video-generation quality on LIBERO-Plus (1,415,986 shared action-replay clips; nine frames with the reconstructed conditioning frame included). Bold: better score; gray: BridgeWAM minus Fast-WAM.}
\label{tab:idm_video_generation}
\begin{tabular}{p{\dimexpr0.22\linewidth-2\tabcolsep\relax}*{5}{>{\centering\arraybackslash}p{\dimexpr0.156\linewidth-2\tabcolsep\relax}}}
\toprule
\textbf{Method}
& \textbf{FID} $\downarrow$
& \textbf{FVD$_9$} $\downarrow$
& \textbf{SSIM} $\uparrow$
& \textbf{LPIPS} $\downarrow$
& \textbf{PSNR} (dB) $\uparrow$ \\
\midrule
\textbf{Fast-WAM IDM}
& 11.3573 & 30.8644 & 0.771549 & 0.202842 & 20.5026 \\
\rowcolor{bridgewamrow!60!white}
\textbf{\sysname{} IDM}
& \makecell[c]{\textbf{11.2093}\\[-2pt]\textcolor{black!55}{\scriptsize$-0.1480$}} & \makecell[c]{\textbf{30.7990}\\[-2pt]\textcolor{black!55}{\scriptsize$-0.0654$}} & \makecell[c]{\textbf{0.773735}\\[-2pt]\textcolor{black!55}{\scriptsize$+0.002186$}} & \makecell[c]{\textbf{0.199942}\\[-2pt]\textcolor{black!55}{\scriptsize$-0.002900$}} & \makecell[c]{\textbf{20.5471}\\[-2pt]\textcolor{black!55}{\scriptsize$+0.0445$}} \\
\bottomrule
\end{tabular}
\end{table}

}

  \subsection{Quantitative Evaluation on Real Robots}
  \label{sec:real_world_results}

Figure~\ref{fig:real_world_results} compares task success and inference latency across the three methods. We deploy \sysname{} on the \textit{PiPER X from SongLing} real-robot platform and evaluate four manipulation tasks with their OOD settings. We use task-specific staged scoring, with final scores computed over completed operation stages; for example, within a single trial, successfully grasping an object and achieving the target each earn credit for their respective stages. For drink pick-and-place, grasping the drink, placing it within the target region, and leaving it upright each earn one point. Scoring criteria are detailed in Appendix~\ref{sec:appendix_real_scoring}. These results show that the LBQ-mediated video--action interface transfers beyond simulation and maintains stable, action-relevant representations under visual variation, observation noise, and execution disturbances during closed-loop manipulation.\looseness=-1

  \subsection{Ablation Studies and Efficiency}
  \label{sec:ablation_results}

  We organize our ablations around four questions concerning joint visual--action adaptation, the readout of LBQs, multimodal conditioning, and the capacity of the Action Expert. We further compare video--action coupling paradigms and inference efficiency.\looseness=-1

  \paragraph{RQ1: Do video representations need to adapt jointly with action learning?}
  We construct a Frozen-Video-DiT variant that freezes the Video Expert from Fast-WAM trained on LIBERO and conditions the Action Expert through LBQs. In Table~\ref{tab:interface_ablation}, this variant achieves average success rates of $85.50\%$ on LIBERO and $25.48\%$ on LIBERO-Plus, below the full model's $98.25\%$ and $72.24\%$, respectively. This result supports adapting video representations jointly with LBQ-conditioned action learning rather than directly relying on fixed, previously learned visual representations.\looseness=-1

  \paragraph{RQ2: Does dedicated LBQ cross-attention outperform MoT-style readout?}
  The MoT-readout variant reads concatenated LBQ and action keys/values through mixed self-attention, while attending to text and state conditions through cross-attention. Its LIBERO average is $97.80\%$, close to the full model, but its LIBERO-Plus average drops to $60.48\%$ (Table~\ref{tab:interface_ablation}). Compared with this configuration for MoT-style readout, the use of dedicated LBQ cross-attention more effectively supports the prediction of actions under the distribution shifts evaluated here.\looseness=-1

  \paragraph{RQ3: Is unified latent conditioning preferable to directly exposing multimodal tokens?}
  The Text-state-LBQ-input variant supplies text, state, and LBQ tokens together as inputs to action cross-attention. Its average success rates on LIBERO and LIBERO-Plus are $97.20\%$ and $59.29\%$, respectively, both below the full model (Table~\ref{tab:interface_ablation}). This result further suggests that the prior information used by the Action Expert can be compressed into a limited LBQ latent space, self-aligned across modalities within that space, and then jointly read and integrated by the action tokens.\looseness=-1

  \paragraph{RQ4: How do Action Expert depth and latent-space capacity affect generalization?}
  We compare Action Experts of different depths while keeping the Video Expert, LBQ count, and training and evaluation protocols fixed. Table~\ref{tab:action_head_depth} shows that scaling the action head does not improve the standard LIBERO score. This pattern supports our hypothesis that excessive action-side capacity can encourage the model to absorb task-irrelevant visual details, weakening its robustness to distribution shifts rather than improving action prediction. We further vary the LBQ sequence length with a two-layer Action Expert and last-layer readout (Table~\ref{tab:lbq_length_ablation}). Within the compact range of 24--36 LBQs, scaling the bridge latent space has little effect on observed task performance, suggesting that a limited latent space can provide sufficient conditioning for action prediction. Meanwhile, the results with 64 LBQs show that further increasing LBQ length does not improve generalization.

  \paragraph{Additional comparison of video--action coupling paradigms.}
  With the same video backbone and nominal two-layer Action Expert configuration, the Joint and IDM variants perform synchronous video--action generation and video generation followed by action generation, respectively. Their LIBERO-Plus averages are $63.41\%$ and $67.45\%$, below the default model's $72.24\%$ (Table~\ref{tab:interface_ablation}). The detailed settings used for each of these variants are provided in Appendix~\ref{sec:appendix_ablation}.

  \paragraph{Inference efficiency.}
  Table~\ref{tab:inference_efficiency} shows that \sysname{} requires $1.460$\,T FLOPs, compared with $2.048$\,T for Fast-WAM and $3.221$\,T for Faster-WAM; the corresponding H20Z inference latencies are $88.05$\,ms, $405.91$\,ms, and $321.16$\,ms, respectively. All three models use \texttt{bfloat16}, an identical cached LIBERO input, a 32-step action horizon, and 10 denoising steps. Latency averages 10 CUDA-event timings after three warm-ups, covering computation from Video DiT layer~0 to the final clean action latent and excluding observation preprocessing, VAE/text encoding, and postprocessing. FLOPs are computed from runtime tensor shapes and include conditioning computation and all action-denoising steps; Appendix~\ref{sec:appendix_efficiency} details the protocol. Together with task performance, these results demonstrate \sysname{}'s balance between execution capability and inference efficiency.

  \paragraph{Video-generation quality under IDM inference.}
  We evaluate \sysname\  IDM and Fast-WAM IDM through zero-shot inference on LIBERO-Plus using 1,415,986 shared action-replay clips. For both models, we assess the complete video clips that are generated by the Video Expert during inference. As shown in Table~\ref{tab:idm_video_generation}, \sysname\  IDM achieves small but consistent improvements across all five metrics. These results indicate that, under the evaluated perturbations, \sysname\  IDM produces task-conditioned visual predictions that more closely match the reference videos. This observation is consistent with our hypothesis that joint visual--action learning through LBQs helps guide the Video Expert toward representations useful to the Action Expert while supporting more accurate prediction of future frames in out-of-distribution scenarios.\looseness=-1

  \subsection{Qualitative Analysis}
  \label{sec:qualitative_results}
  \begin{figure}[t]
      \centering
      \includegraphics[width=\linewidth,trim={0cm 3cm 0cm 1cm}]{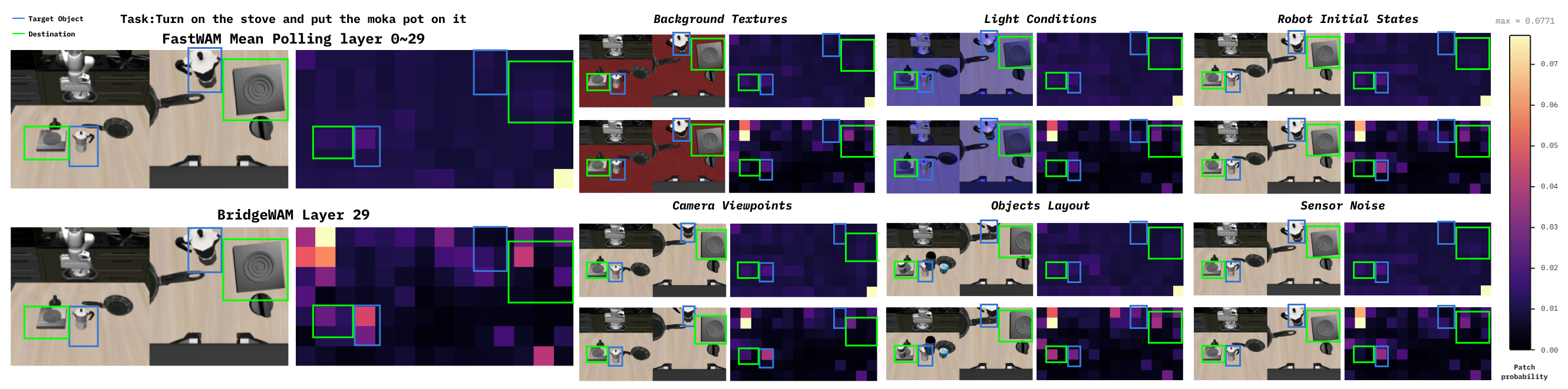}
      \caption{\textbf{Action-conditioned visual routing maps.}
      The top row shows Fast-WAM attention averaged across Video-DiT layers,
      while the bottom row shows the LBQ-mediated routing at the final Video-DiT layer of \sysname{}.
      The columns show the models under variations in background texture, lighting, camera viewpoint,
      robot initial state, object layout, and sensor noise.\looseness=-1}
      \label{fig:action_conditioned_routing}
  \end{figure}
  \paragraph{Action-Conditioned Visual Routing.}
  To visualize which visual information is preferentially accessed by the Action Expert, we construct an action-conditioned visual-routing map. For \sysname{}, we use the Action Expert's attention scores over the latent bridge queries as query-importance weights and apply them to the corresponding LBQ-to-video-patch attention maps from the Video Expert. For Fast-WAM, we directly aggregate the action-to-visual-token attention across all Video-DiT layers. The resulting patch-space maps reveal the visual information preferentially routed toward action prediction; they should be interpreted as routing visualizations rather than strict causal attributions. Figure~\ref{fig:action_conditioned_routing} shows representative routing maps for the two models under changes in background texture, lighting, camera viewpoint, robot initial state, object layout, and sensor noise. Experimental results show that the BoE not only enables the Action Expert to focus its attention on visual features that are more relevant to the current task, but also improves robustness to perturbations and enhances the model’s generalization performance across the evaluated visual changes. The layer selection follows each model’s visual-conditioning interface: Fast-WAM reads video features through layer-wise MoT connections, so we average attention across layers to summarize its overall attention preferences when accessing visual information at different depths; \sysname\  conditions its Action Expert on the final LBQ representations, so we visualize the action-weighted LBQ-to-patch attention at this interface. These maps characterize architecture-specific routing preferences rather than provide a matched-layer comparison.\looseness=-1

\section{Conclusion}
\label{sec:conclusion}

We introduced Bridge-of-Experts (BoE), instantiated as \sysname{}, to address two limitations of layer-wise MoT coupling: direct exposure of dense visual features to action layers and the requirement for matched expert depths. BoE uses a compact, learnable LBQ interface to aggregate multimodal information, enabling selective retrieval by the Action Expert while allowing action supervision to shape video representations. Decoupling expert depths further enables lightweight action decoding without reducing the Video Expert's depth. Together, these designs alleviate competition between task-relevant information and redundant visual details and reduce action-side computation, supporting a favorable balance between task performance, generalization, and inference efficiency in the evaluated settings.\looseness=-1

\textbf{Limitations.} During training on real-world robot tasks, however, we observed that this paradigm struggles to fit more complex, long-horizon manipulation tasks. \sysname{} also retains instability observed in its Fast-WAM foundation.\looseness=-1

\textbf{Future Outlook.} Future work will focus on improving learning for long-horizon manipulation and constructing a more stable, capacity-limited latent space. We will investigate how the bridge can preserve task-relevant information over longer temporal spans and provide more consistent conditioning for action prediction while maintaining the efficiency of lightweight action decoding.

\subsection*{AI Use Statement}
We used large language models (LLMs) to polish the language of author-written
text for clarity, fluency, and readability. For example, when preparing the
Related Work section, the authors first read and organized relevant papers
and manually added their citations to the bibliography. Based on their
understanding of the field's technical development, the authors then drafted
the discussion of how prior work relates to the core technical ideas and how
these ideas evolved. The resulting author-written paragraphs were subsequently
provided to LLMs for language polishing. The authors take responsibility for
the final manuscript, including its claims, citations, and reported results.

\subsection*{Reproducibility Statement}
We provide a link to the project homepage at the end of the abstract and
a link to our Git repository at the beginning of the appendix.
To support reproducibility, we describe the model architecture and training
objectives in Section~3, and the experimental setup and simulation evaluation
protocols in Sections~\ref{sec:experimental_setup}
and~\ref{sec:simulation_results}.
Appendix~\ref{sec:appendix_initialization} describes action-head initialization,
and Appendix~\ref{sec:appendix_ablation} specifies the ablation variants.
The staged scoring approach for real-robot experiments is described in
Appendix~\ref{sec:appendix_real_scoring}.
Appendix~\ref{sec:appendix_efficiency} documents the hardware, numerical
precision, timing boundaries, and FLOPs accounting used for efficiency
measurements.

\bibliography{references}
\bibliographystyle{bridgewam}

\clearpage
\appendix
\section{Additional Experimental Details}
\label{sec:appendix}

\noindent\textbf{Code repository:}
\url{https://github.com/jxfang04/BridgeWAM-code.git}.

This appendix records implementation and evaluation details that complement
the main text. Unless stated otherwise, ablations use the same data splits,
optimizer, numerical precision, denoising schedule, and checkpoint-selection
rule of the full \sysname{} model throughout these comparisons.

\subsection{Experimental Design Details}
\label{sec:appendix_initialization}

In default training and inference, \sysname{} combines
pretrained backbone weights with randomly initialized task-specific modules.
We detail the initialization of each component below.\looseness=-1

\paragraph{Pretrained visual and text components.}
The Video Expert uses the pretrained Wan2.2-TI2V-5B Video DiT, with 30 layers
and a hidden dimension of 3072. We load the corresponding pretrained weights
for its attention and feed-forward networks (FFNs), text and timestep
projections, and video output head. The VAE encoder and decoder also use
pretrained Wan2.2 weights and remain frozen throughout training. The text conditioning is provided by UMT5-XXL. By default, the training process reads precomputed text embeddings without loading a separate encoder for text; when enabled, the text
encoder is initialized with its pretrained weights.\looseness=-1

\paragraph{Action Expert initialization.}
The Action Expert is a two-layer Action DiT with a hidden dimension of 1024,
using an alternating cross--self configuration that performs cross-attention
followed by self-attention. Its two layers are initialized from the
corresponding parameters of layers 0 and 29, respectively, in a
dimension-adapted, Wan-derived Action DiT checkpoint, including attention,
FFN, and modulation parameters. The action timestep encoder
($256\!\rightarrow\!1024\!\rightarrow\!1024$), timestep modulation projection
($1024\!\rightarrow\!6144$), and text projection
($4096\!\rightarrow\!1024\!\rightarrow\!1024$) are loaded from the same
preprocessed checkpoint. The projection for action text is retained in the model but is not used in the forward pass under \texttt{lbq\_only} conditioning.\looseness=-1

\paragraph{Newly initialized projections.}
The action input encoder ($7\!\rightarrow\!1024$), action output head
($1024\!\rightarrow\!7$), two-layer LBQ-to-action projector
($3072\!\rightarrow\!1024\!\rightarrow\!1024$), and proprioceptive encoder
($8\!\rightarrow\!4096$) use the default PyTorch \texttt{Linear} random
initialization. Each linear layer's weights and biases are sampled from a
uniform distribution whose bounds depend on that layer's input dimension.

\paragraph{Learnable embeddings and deterministic components.}
The action positional embeddings have shape $[1024,1024]$ and are initialized
from a zero-mean Gaussian distribution with a standard deviation of $0.02$.
The first dimension specifies the maximum number of supported positions,
rather than the executed action horizon. The 32 LBQs form a learnable token
bank of shape $[1,32,3072]$, initialized from a zero-mean Gaussian distribution
with a standard deviation of $1/\sqrt{3072}\approx0.01804$. These embeddings
are persistent trainable parameters and are not resampled at each forward
pass. RoPE, attention masks, and the flow-matching scheduler are constructed
from deterministic rules and configuration settings; they contain no
trainable weights requiring random initialization.

\subsection{Ablation Protocol and Variant Definitions}
\label{sec:appendix_ablation}

We study action-head capacity, three interface ablations, and two alternative
video--action generation schemes. The interface and generation variants in
Table~\ref{tab:interface_ablation} retain the default backbone and nominal
two-layer Action Expert, with the specific changes described in the paragraphs below.\looseness=-1

\paragraph{Action-head depth.}
The depth study varies the Action Expert across 4, 8, 16, and 30 layers while
keeping the Video Expert, LBQ count, training schedule, and evaluation
protocol fixed. The 30-layer variant uses a self-cross baseline action head
whose nominal depth matches the Video Expert.

\paragraph{Frozen-Video-DiT: adapting visual features for control.}
This variant tests whether the Video Expert must adapt its features to
LBQ-mediated action learning. It retains the default training and inference
architecture but initializes the Video DiT with Fast-WAM weights trained on
LIBERO and freezes these weights. The Action Expert reads features produced
by this frozen backbone through the LBQs. Its LIBERO-Plus average drops to
$25.48\%$, compared with $72.24\%$ for the full model, supporting the need to
adapt video representations jointly with action learning in this setting.

\paragraph{MoT-readout: changing how actions read LBQs.}
This variant tests whether dedicated LBQ cross-attention improves the
Action Expert's use of task-relevant visual features. During training and inference, mixed self-attention concatenates LBQ and action keys/values, while action queries attend to text--state keys/values in
cross-attention. The LIBERO-Plus average falls to $60.48\%$, indicating that
this readout is less effective than the dedicated LBQ cross-attention used
by the full model.

\paragraph{Text-state-LBQ-input: mixing conditioning sources.}
This variant tests whether conditioning through a unified LBQ space is
preferable to providing the Action Expert with direct access to additional modalities.
During training and inference, action queries in cross-attention read
concatenated text, state, and LBQ keys/values. Its LIBERO-Plus average of
$59.29\%$ indicates reduced robustness to the evaluated perturbations
relative to conditioning exclusively through LBQs in the full model.

\paragraph{Joint: synchronous generation through LBQs.}
Joint propagates video, LBQ, and action states together in each denoising
forward pass. At Video DiT layers 0--28, the same Action DiT layer-0
cross-attention block is repeatedly applied, reading the LBQs available at
the start of each stage. Video queries attend to video and LBQ keys/values,
whereas LBQ queries attend to video, LBQ, and action keys/values. At Video
DiT layer 29, Action DiT layer 1 performs pure action self-attention once.
Text--state conditioning enters only the Video/LBQ side. We stop gradients
through the action keys/values read by LBQs, so the Action Expert is updated
only by the action loss, while the Video Expert and LBQs receive both video
and action supervision. At inference, video and action latents are denoised
synchronously; the standard \texttt{infer\_action()} path returns actions
without executing the VAE decoder.\looseness=-1

\paragraph{IDM: video generation followed by action generation.}
IDM first completes multi-step video denoising at inference, then conditions
action generation on 32 LBQs representing the full video. After the action
projector, these LBQs supply the sole keys/values for Action DiT layer-0
cross-attention; layer 1 performs action-only self-attention. The same LBQ context is reused throughout the entire sequence of action denoising steps. Training instead uses
two Video/LBQ forward passes: one predicts video and incurs the video loss;
the other extracts action-conditioning LBQs from complete ground-truth video
latents, which are either clean or noise-corrupted, with a corruption
probability of $0.5$. Thus, training does not first sample a complete video
through iterative denoising. The current inference implementation also runs
the VAE decoder. Table~\ref{tab:idm_inference_strategies} distinguishes the
three inference strategies.
The $67.45\%$ LIBERO-Plus result in Table~\ref{tab:interface_ablation} uses
a full-frame prefill at $t=0$ on the final video latent $x_0$ to obtain the LBQs used as conditioning for action denoising.

\begin{table}[H]
    \centering
    \caption{IDM inference strategies. Prefill denotes an additional
    full-frame Video/LBQ forward pass at $t=0$ on the final denoised video
    latent $x_0$. Frame access refers to the video frames visible to LBQ
    queries during video denoising.}
    \label{tab:idm_inference_strategies}

    \setlength{\tabcolsep}{4pt}
    \renewcommand{\arraystretch}{1.15}
    \begin{tabular}{@{}p{0.23\linewidth}p{0.22\linewidth}p{0.47\linewidth}@{}}
        \toprule
        Strategy & LBQ frame access & LBQs used for action conditioning \\
        \midrule
        A (original IDM) & First frame only
            & Extra full-frame prefill on final $x_0$. \\
        B1 (full-video default) & All frames
            & Final-layer LBQs from the last video denoising forward pass;
              no extra prefill. \\
        B2 & All frames
            & Extra full-frame prefill on final $x_0$. \\
        \bottomrule
    \end{tabular}
\end{table}

B1 uses hidden states computed before the final video latent update, so its
LBQs are not an encoding of the final $x_0$. B1 and B2 can share the same
full-video checkpoint. Joint and IDM therefore test synchronous interaction
and sequential video-conditioned action generation, respectively, while
both preserve LBQ-mediated communication without direct video-to-action
attention.\looseness=-1

\paragraph{Interpretation.}
Together, these ablations support learning the external conditioning needed
for the evaluated tasks within the Video Expert and consolidating it into a
fixed-size LBQ space for task-relevant action readout. One possible
explanation for the weaker MoT-readout and text-state-LBQ-input results is
that concatenating LBQ keys/values with tokens from other representation
spaces introduces distribution mismatch and attention competition, weakening
selective access to useful visual features and increasing reliance on
irrelevant information. The success-rate comparisons support the unified
interface but do not establish this proposed mechanism causally.\looseness=-1

\clearpage
\subsection{Detailed Results on Simulation Benchmarks}
\label{sec:appendix_simulation_results}

\paragraph{LIBERO-Plus.}
Table~\ref{tab:appendix_libero_plus_detailed} provides detailed results for
\sysname{} across four suites, seven perturbation categories, and the
recorded difficulty levels. The evaluation covers 10,030 task variants with
one episode per variant; all episodes are complete. Pooling the 7,246
successful episodes yields an overall success rate of $72.24\%$. Each pooled
entry is computed as total successes divided by total episodes in the
corresponding group. The macro column instead gives equal weight to the four
suite-level rates; its overall value is $72.39\%$. We distinguish these
aggregation rules because the suites and perturbation categories contain
different numbers of episodes.\looseness=-1

\begin{table}[H]
\centering

\caption{\textbf{Detailed LIBERO-Plus results for \sysname{}.} Success rates are
reported in percent (higher is better). Panel (a) separates suite-wise rates,
their unweighted macro average, and success rates computed by pooling episodes. Panel (b)
groups episodes by the recorded difficulty level. All pooled totals include the episodes for which no difficulty level is specified.}
\label{tab:appendix_libero_plus_detailed}
\setlength{\tabcolsep}{3pt}
\renewcommand{\arraystretch}{1.12}
\begin{tabular}{@{}lrrrrrrr@{}}
\toprule
\multicolumn{8}{@{}l}{\textbf{(a) Results by suite and perturbation category}} \\
\midrule
Perturbation & Spatial & Object & Goal & Long & Macro & Pooled & Episodes \\
\midrule
Camera viewpoint & 65.16 & 83.59 & 30.64 & 46.78 & 56.54 & 56.10 & 1,599 \\
Robot initial state & 60.29 & 62.81 & 51.83 & 66.92 & 60.46 & 60.39 & 1,550 \\
Language instruction & 86.15 & 95.76 & 69.76 & 88.77 & 85.11 & 84.65 & 1,537 \\
Lighting & 97.26 & 95.29 & 94.98 & 81.39 & 92.23 & 92.38 & 1,142 \\
Background texture & 84.50 & 86.29 & 64.41 & 41.52 & 69.18 & 68.12 & 1,076 \\
Sensor noise & 76.35 & 91.94 & 60.69 & 65.26 & 73.56 & 73.64 & 1,601 \\
Object layout & 78.70 & 79.90 & 61.88 & 82.37 & 75.71 & 75.08 & 1,525 \\
\midrule
\textbf{All perturbations} & 77.64 & 84.47 & 60.29 & 67.17 & 72.39 & \textbf{72.24} & 10,030 \\
Episodes & 2,402 & 2,518 & 2,591 & 2,519 & -- & -- & 10,030 \\
\midrule
\multicolumn{8}{@{}l}{\textbf{(b) Results by difficulty level and perturbation category}} \\
\midrule
Perturbation & Level 1 & Level 2 & Level 3 & Level 4 & Level 5 & Unspec. & Pooled \\
\midrule
Camera viewpoint & 70.85 & 60.28 & 56.92 & 57.89 & 45.59 & -- & 56.10 \\
Robot initial state & 87.50 & 82.19 & 64.97 & 45.26 & 30.48 & -- & 60.39 \\
Language instruction & 90.85 & 93.57 & 92.28 & 85.67 & 64.62 & -- & 84.65 \\
Lighting & 98.20 & 94.42 & 94.67 & 91.74 & 83.26 & 97.52 & 92.38 \\
Background texture & 87.55 & 78.23 & 69.58 & 55.21 & 30.14 & -- & 68.12 \\
Sensor noise & 81.36 & 84.60 & 80.00 & 70.50 & 42.32 & -- & 73.64 \\
Object layout & 86.49 & 86.75 & 78.42 & 71.60 & 48.44 & -- & 75.08 \\
\midrule
\textbf{All perturbations} & 86.01 & 81.83 & 75.41 & 69.35 & 49.21 & 97.52 & \textbf{72.24} \\
Episodes & 1,644 & 2,202 & 2,094 & 1,886 & 2,083 & 121 & 10,030 \\
\bottomrule
\end{tabular}
\par\vspace{3pt}
\parbox{\linewidth}{ Long denotes the \texttt{libero\_10} suite.
Unspec. denotes the 121 lighting episodes for which no difficulty level was recorded
(118 successes); -- indicates cells in which there are no episodes.}
\end{table}

The breakdown reveals substantial variation across perturbations. Pooled
success is highest under lighting changes ($92.38\%$) and lowest under camera
viewpoint changes ($56.10\%$). At difficulty level 5, background texture and
robot initial-state perturbations yield success rates of $30.14\%$ and
$30.48\%$, respectively, highlighting challenging conditions that the
overall average obscures.

\clearpage
\paragraph{LIBERO-Pro.}
Table~\ref{tab:appendix_libero_pro_detailed} reports \sysname{} results for
five perturbation groups across the four suites. Each of the 20
suite--perturbation combinations contains ten tasks evaluated with 50 trials
each, giving 200 task variants and 10,000 episodes. All planned variants are
complete, with 5,372 successful episodes and an overall success rate of
$53.72\%$. Because each combination contains 500 episodes, the pooled overall
rate also equals the unweighted mean across the 20 combinations. The
evaluation uses the \texttt{main\_table} environment variant together with
the official non-environment variants, as specified by the generation
manifest; seed 28 is used for initial-state generation.

\begin{table}[H]
\centering

\caption{\textbf{Detailed LIBERO-Pro results for \sysname{}.} Success rates
are reported in percent (higher is better). Each suite--perturbation cell
contains ten tasks with 50 trials each. Pooled rates combine all episodes
in the corresponding row; success and episode counts cover all four suites.}
\label{tab:appendix_libero_pro_detailed}
\setlength{\tabcolsep}{4pt}
\renewcommand{\arraystretch}{1.15}
\begin{tabular}{@{}lrrrrrrr@{}}
\toprule
Perturbation & Spatial & Object & Goal & Long & Pooled & Successes & Episodes \\
\midrule
Language & 96.20 & 99.20 & 97.80 & 97.20 & 97.60 & 1,952 & 2,000 \\
Object & 96.60 & 91.80 & 83.00 & 56.60 & 82.00 & 1,640 & 2,000 \\
Swap & 0.00 & 0.00 & 4.60 & 0.80 & 1.35 & 27 & 2,000 \\
Task & 0.00 & 0.00 & 10.00 & 9.80 & 4.95 & 99 & 2,000 \\
Environment & 96.40 & 84.60 & 96.80 & 53.00 & 82.70 & 1,654 & 2,000 \\
\midrule
\textbf{All perturbations} & 57.84 & 55.12 & 58.44 & 43.48 & \textbf{53.72} & 5,372 & 10,000 \\
Episodes & 2,500 & 2,500 & 2,500 & 2,500 & -- & -- & 10,000 \\
\bottomrule
\end{tabular}
\par\vspace{3pt}
\parbox{\linewidth}{ Long denotes \texttt{libero\_10}.
The names of the perturbation groups correspond to the labels used in the evaluation \texttt{lan},
\texttt{object}, \texttt{swap}, \texttt{task}, and \texttt{env}, respectively.}
\end{table}

Success rates vary sharply across perturbation groups: language,
environment, and object perturbations yield $97.60\%$, $82.70\%$, and
$82.00\%$, respectively, whereas the recorded swap and task groups yield
$1.35\%$ and $4.95\%$. These low scores identify limitations obscured by the aggregate result.

\clearpage
\paragraph{RoboTwin 2.0.}
Table~\ref{tab:robotwin_finegrained_baselines_full} reports the complete results for
all 50 tasks under the clean and randomized settings. We evaluate the
checkpoint at training step 29,355 using the \texttt{unseen} instruction
setting, with 100 episodes per task and condition, for 10,000 episodes in
total. \sysname{} succeeds in 4,492 of 5,000 clean episodes ($89.84\%$) and
4,511 of 5,000 randomized episodes ($90.22\%$), yielding $90.03\%$ when both
settings are pooled. Since the number of episodes is the same for each task, the
per-task mean and pooled success rate coincide within each setting.\looseness=-1

\providecommand{\sysname}{\textsc{LQ-WAM}}
\providecommand{\cmark}{\ding{51}}
\providecommand{\xmark}{\ding{55}}

\begin{table*}[t]
\centering
\caption{Fine-grained task comparison on RoboTwin 2.0 Simulation
(Clean vs. Randomized).}
\label{tab:robotwin_finegrained_baselines_full}
\scriptsize
\resizebox{\textwidth}{!}{
\begin{tabular}{l|cc|cc|cc|cc|cc|cc|cc|cc|cc}
\toprule
\textbf{Task}
& \multicolumn{2}{c|}{$\pi_{0}$}
& \multicolumn{2}{c|}{$\pi_{0.5}$}
& \multicolumn{2}{c|}{Motus}
& \multicolumn{2}{c|}{Motus from WAN2.2}
& \multicolumn{2}{c|}{LingBot-VA}
& \multicolumn{2}{c|}{LingBot-VA from WAN2.2}
& \multicolumn{2}{c|}{Fast-WAM}
& \multicolumn{2}{c|}{Faster-WAM}
& \multicolumn{2}{c}{BridgeWAM (Ours)} \\
& \multicolumn{2}{c|}{\cite{black2024pi0}}
& \multicolumn{2}{c|}{\cite{pi05}}
& \multicolumn{2}{c|}{\cite{bi2025motus}}
& \multicolumn{2}{c|}{\cite{bi2025motus}}
& \multicolumn{2}{c|}{\cite{li2026lingbotva}}
& \multicolumn{2}{c|}{\cite{li2026lingbotva}}
& \multicolumn{2}{c|}{\cite{yuan2026fastwam}}
& \multicolumn{2}{c|}{\cite{zhao2026faster}}
& \multicolumn{2}{c}{} \\
& C & R & C & R & C & R & C & R & C & R
& C & R & C & R & C & R & C & R \\
\midrule

Adjust Bottle
& 99 & 95 & 100 & 99 & 89 & 93 & 99 & 97 & 90 & 94 & -- & --
& 100 & 100 & 100 & 99 & 100 & 99 \\

Beat Block Hammer
& 79 & 84 & 96 & 93 & 95 & 88 & 88 & 90 & 96 & 98 & -- & --
& 99 & 97 & 97 & 98 & 99 & 91 \\

Blocks Ranking RGB
& 80 & 63 & 92 & 85 & 99 & 97 & 92 & 88 & 99 & 98 & -- & --
& 100 & 100 & 100 & 100 & 99 & 99 \\

Blocks Ranking Size
& 14 & 5 & 49 & 26 & 75 & 63 & 38 & 50 & 94 & 96 & -- & --
& 94 & 98 & 74 & 93 & 87 & 94 \\

Click Alarmclock
& 77 & 68 & 98 & 89 & 100 & 100 & 100 & 99 & 99 & 100 & -- & --
& 100 & 100 & 99 & 100 & 97 & 99 \\

Click Bell
& 71 & 48 & 99 & 66 & 100 & 100 & 100 & 100 & 100 & 100 & -- & --
& 100 & 100 & 100 & 100 & 99 & 99 \\

Dump Bin Bigbin
& 88 & 83 & 92 & 97 & 95 & 91 & 94 & 96 & 89 & 96 & -- & --
& 97 & 96 & 97 & 98 & 98 & 97 \\

Grab Roller
& 98 & 94 & 100 & 100 & 100 & 100 & 100 & 100 & 100 & 100 & -- & --
& 100 & 100 & 100 & 100 & 100 & 100 \\

Handover Block
& 47 & 31 & 66 & 57 & 86 & 73 & 34 & 15 & 99 & 78 & -- & --
& 95 & 81 & 93 & 88 & 83 & 80 \\

Handover Mic
& 97 & 97 & 98 & 97 & 78 & 63 & 98 & 95 & 94 & 96 & -- & --
& 99 & 100 & 100 & 99 & 100 & 100 \\

Hanging Mug
& 14 & 11 & 18 & 17 & 38 & 38 & 14 & 10 & 40 & 28 & -- & --
& 58 & 62 & 61 & 44 & 32 & 31 \\

Lift Pot
& 80 & 72 & 96 & 85 & 96 & 99 & 90 & 87 & 100 & 99 & -- & --
& 100 & 100 & 100 & 100 & 100 & 100 \\

Move Can Pot
& 68 & 48 & 51 & 55 & 34 & 74 & 43 & 53 & 94 & 97 & -- & --
& 90 & 88 & 96 & 98 & 78 & 98 \\

Move Pillbottle Pad
& 67 & 46 & 84 & 61 & 93 & 96 & 83 & 83 & 99 & 99 & -- & --
& 100 & 99 & 98 & 99 & 99 & 100 \\

Move Playingcard Away
& 74 & 65 & 96 & 84 & 100 & 96 & 50 & 47 & 100 & 99 & -- & --
& 100 & 100 & 100 & 100 & 98 & 100 \\

Move Stapler Pad
& 41 & 24 & 56 & 42 & 83 & 85 & 49 & 37 & 91 & 79 & -- & --
& 77 & 64 & 82 & 74 & 80 & 67 \\

Open Laptop
& 71 & 81 & 90 & 96 & 95 & 91 & 89 & 89 & 92 & 94 & -- & --
& 98 & 100 & 95 & 99 & 99 & 100 \\

Open Microwave
& 4 & 32 & 34 & 77 & 95 & 91 & 83 & 82 & 82 & 86 & -- & --
& 62 & 45 & 77 & 77 & 50 & 56 \\

Pick Diverse Bottles
& 69 & 31 & 81 & 71 & 90 & 91 & 53 & 62 & 89 & 82 & -- & --
& 80 & 85 & 90 & 89 & 70 & 74 \\

Pick Dual Bottles
& 59 & 37 & 93 & 63 & 96 & 90 & 58 & 68 & 100 & 99 & -- & --
& 100 & 96 & 98 & 98 & 89 & 87 \\

Place A2B Left
& 43 & 47 & 87 & 82 & 88 & 79 & 78 & 79 & 97 & 93 & -- & --
& 95 & 93 & 96 & 94 & 94 & 95 \\

Place A2B Right
& 39 & 34 & 87 & 84 & 91 & 87 & 86 & 83 & 97 & 95 & -- & --
& 93 & 99 & 94 & 93 & 95 & 95 \\

Place Bread Basket
& 62 & 46 & 77 & 64 & 91 & 94 & 73 & 83 & 97 & 95 & -- & --
& 91 & 93 & 89 & 96 & 94 & 97 \\

Place Bread Skillet
& 66 & 49 & 85 & 66 & 86 & 83 & 71 & 71 & 95 & 90 & -- & --
& 90 & 93 & 93 & 87 & 89 & 92 \\

Place Burger Fries
& 81 & 76 & 94 & 87 & 98 & 98 & 95 & 90 & 97 & 95 & -- & --
& 96 & 99 & 98 & 97 & 98 & 98 \\

Place Can Basket
& 55 & 46 & 62 & 62 & 81 & 76 & 46 & 62 & 81 & 84 & -- & --
& 71 & 69 & 76 & 73 & 78 & 79 \\

Place Cans Plasticbox
& 63 & 45 & 94 & 84 & 98 & 94 & 96 & 99 & 100 & 99 & -- & --
& 99 & 96 & 99 & 94 & 96 & 92 \\

Place Container Plate
& 97 & 92 & 99 & 95 & 98 & 99 & 97 & 100 & 99 & 97 & -- & --
& 96 & 100 & 98 & 97 & 99 & 100 \\

Place Dual Shoes
& 59 & 51 & 75 & 75 & 93 & 87 & 78 & 80 & 94 & 89 & -- & --
& 94 & 88 & 91 & 88 & 85 & 87 \\

Place Empty Cup
& 91 & 85 & 100 & 99 & 99 & 98 & 97 & 97 & 100 & 100 & -- & --
& 100 & 100 & 100 & 100 & 98 & 100 \\

Place Fan
& 66 & 71 & 87 & 85 & 91 & 87 & 77 & 85 & 99 & 93 & -- & --
& 96 & 96 & 98 & 93 & 99 & 92 \\

Place Mouse Pad
& 20 & 20 & 60 & 39 & 66 & 68 & 62 & 68 & 93 & 96 & -- & --
& 83 & 89 & 86 & 94 & 86 & 86 \\

Place Object Basket
& 67 & 70 & 80 & 76 & 81 & 87 & 74 & 75 & 91 & 88 & -- & --
& 89 & 88 & 80 & 80 & 81 & 78 \\

Place Object Scale
& 57 & 52 & 86 & 80 & 88 & 85 & 84 & 83 & 96 & 95 & -- & --
& 90 & 97 & 92 & 97 & 91 & 96 \\

Place Object Stand
& 82 & 68 & 91 & 85 & 98 & 97 & 91 & 93 & 99 & 96 & -- & --
& 90 & 94 & 95 & 96 & 96 & 98 \\

Place Phone Stand
& 49 & 53 & 81 & 81 & 87 & 86 & 80 & 78 & 97 & 97 & -- & --
& 97 & 99 & 100 & 96 & 98 & 95 \\

Place Shoe
& 76 & 76 & 92 & 93 & 99 & 97 & 95 & 92 & 98 & 98 & -- & --
& 96 & 99 & 97 & 98 & 97 & 96 \\

Press Stapler
& 44 & 37 & 87 & 83 & 93 & 98 & 97 & 94 & 85 & 82 & -- & --
& 90 & 97 & 90 & 92 & 91 & 94 \\

Put Bottles Dustbin
& 65 & 56 & 84 & 79 & 81 & 79 & 36 & 33 & 87 & 91 & -- & --
& 95 & 90 & 94 & 91 & 89 & 92 \\

Put Object Cabinet
& 73 & 60 & 80 & 79 & 88 & 71 & 84 & 64 & 85 & 87 & -- & --
& 94 & 89 & 87 & 89 & 94 & 83 \\

Rotate QRcode
& 74 & 70 & 89 & 87 & 89 & 73 & 80 & 60 & 96 & 91 & -- & --
& 93 & 89 & 94 & 87 & 90 & 84 \\

Scan Object
& 55 & 42 & 72 & 65 & 67 & 66 & 42 & 50 & 96 & 91 & -- & --
& 89 & 92 & 91 & 90 & 88 & 88 \\

Shake Bottle Horizontally
& 98 & 92 & 99 & 99 & 100 & 98 & 100 & 97 & 100 & 99 & -- & --
& 100 & 100 & 100 & 100 & 100 & 100 \\

Shake Bottle
& 94 & 91 & 99 & 97 & 100 & 97 & 100 & 96 & 100 & 97 & -- & --
& 100 & 100 & 100 & 100 & 100 & 100 \\

Stack Blocks Three
& 72 & 52 & 91 & 76 & 91 & 95 & 71 & 76 & 99 & 98 & -- & --
& 95 & 97 & 98 & 96 & 91 & 93 \\

Stack Blocks Two
& 93 & 79 & 97 & 100 & 100 & 98 & 96 & 94 & 100 & 98 & -- & --
& 100 & 100 & 100 & 100 & 100 & 99 \\

Stack Bowls Three
& 77 & 75 & 77 & 71 & 79 & 87 & 90 & 74 & 86 & 83 & -- & --
& 80 & 81 & 83 & 78 & 88 & 90 \\

Stack Bowls Two
& 94 & 95 & 95 & 96 & 98 & 98 & 98 & 98 & 94 & 98 & -- & --
& 92 & 98 & 98 & 96 & 99 & 97 \\

Stamp Seal
& 46 & 33 & 79 & 55 & 93 & 92 & 80 & 88 & 96 & 97 & -- & --
& 90 & 94 & 90 & 92 & 76 & 78 \\

Turn Switch
& 41 & 42 & 62 & 54 & 84 & 78 & 69 & 60 & 44 & 45 & -- & --
& 61 & 59 & 75 & 76 & 55 & 66 \\

\midrule

\textbf{Average (\%)}
& 65.92 & 58.40
& 82.74 & 76.76
& 88.66 & 87.02
& 77.56 & 77.00
& 92.90 & 91.50
& 80.60 & --
& 91.88 & 91.78
& 92.78 & 92.26
& \textbf{89.84} & \textbf{90.22} \\

\bottomrule
\end{tabular}
}
\end{table*}

\sysname{} achieves $100\%$ success in both settings on \textit{Grab Roller},
\textit{Handover Mic}, and \textit{Lift Pot}. Performance is lower on
\textit{Hanging Mug} ($32\%$ clean; $31\%$ randomized) and
\textit{Open Microwave} ($50\%$ clean; $56\%$ randomized), showing that
substantial differences between tasks remain despite the high aggregate
success rates.\looseness=-1

\clearpage
\subsection{Real-Robot Scoring Protocol}
\label{sec:appendix_real_scoring}

We use task-specific staged scoring to record partial completion of the four
manipulation tasks on the \textit{PiPER X from SongLing}, including their
corresponding OOD settings in Fig.~\ref{fig:real_world_results}.
Table~\ref{tab:real_robot_scoring} defines the scoring criteria. Each criterion
contributes one point if satisfied during a trial and zero otherwise; the score for each trial is the sum of these points. Closing the laptop has a single
completion criterion, while the three pick-and-place tasks separately assess
grasping, transport, or placement, with maximum scores of 3, 3, and 4 for the
drink, pen, and mouse tasks, respectively.\looseness=-1

\begin{table}[H]
    \centering
    \caption{\textbf{Task-specific scoring criteria for real-robot evaluation.}
    Each of the numbered criteria contributes one point. Max. denotes the maximum score that can be obtained per trial.}
    \label{tab:real_robot_scoring}

    \setlength{\tabcolsep}{4pt}
    \renewcommand{\arraystretch}{1.15}
    \begin{tabular}{@{}p{0.22\linewidth}p{0.66\linewidth}r@{}}
        \toprule
        Task & One-point criteria & Max. \\
        \midrule
        Close the laptop &
        1. Fully close the laptop lid. & 1 \\
        \addlinespace[5pt]
        Pick and place a drink &
        1. Grasp and lift the drink.\newline
        2. Place the drink within the coaster region.\newline
        3. Leave the drink stably upright after release. & 3 \\
        \addlinespace[5pt]
        Pick and place a pen &
        1. Grasp and lift the pen.\newline
        2. Bring the lifted pen to the vicinity of the pen holder.\newline
        3. Place the pen inside the holder. & 3 \\
        \addlinespace[5pt]
        Pick and place a mouse &
        1. Grasp and lift the mouse.\newline
        2. Maintain a secure grasp during transport without the mouse slipping out.\newline
        3. Place the mouse within the mouse-pad region.\newline
        4. Set the mouse down stably rather than dropping it. & 4 \\
        \bottomrule
    \end{tabular}
\end{table}

\clearpage
\subsection{Frame-by-Frame Comparisons on Real Robots}
\label{sec:appendix_real_sequences}

\begin{figure}[H]
    \centering
    \includegraphics[width=\linewidth]{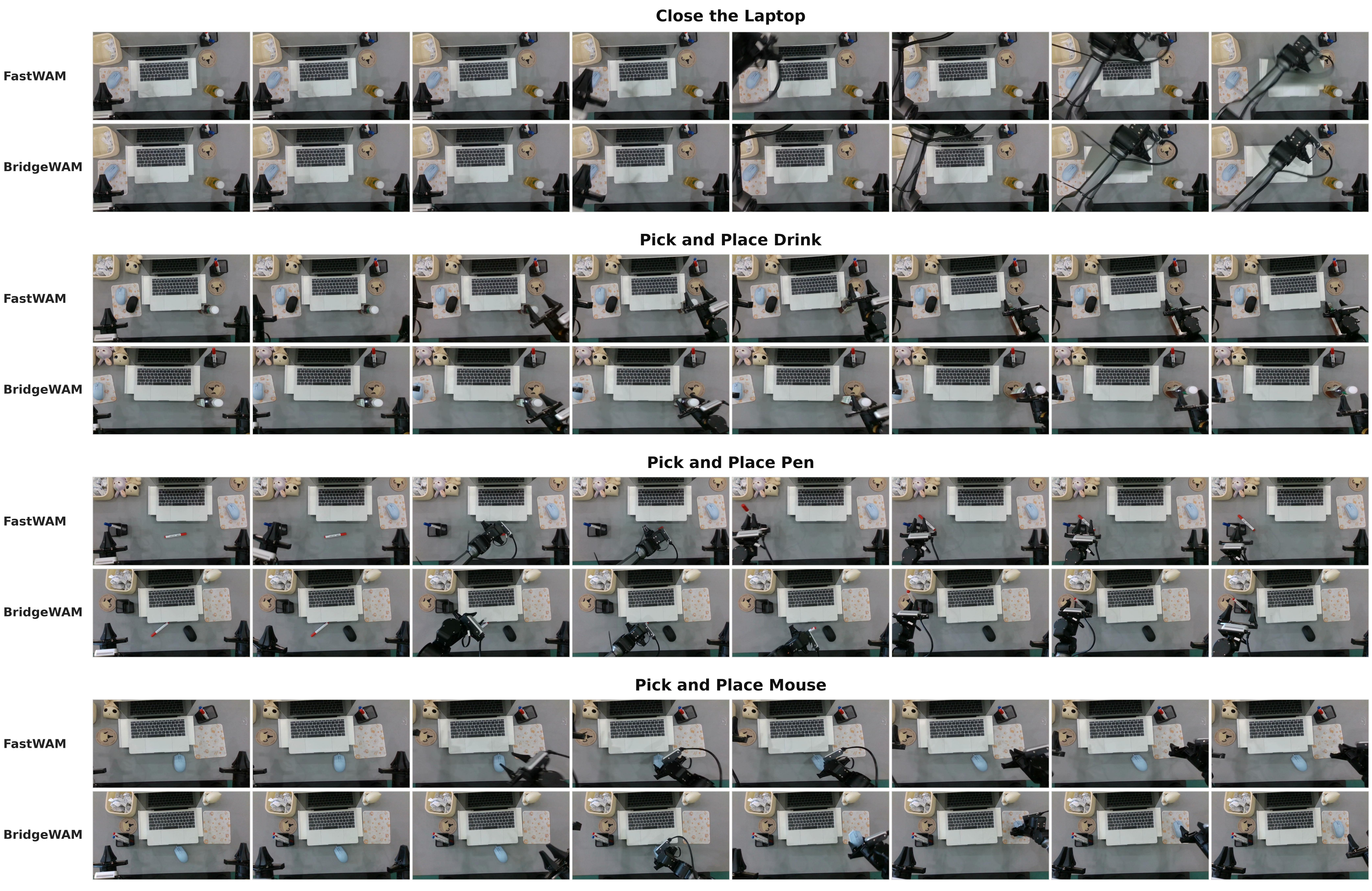}
    \caption{\textbf{Frame-by-frame comparisons on real-robot tasks.}
    The four panels show closing a laptop and picking and placing a drink,
    a pen, and a mouse. Each panel compares Fast-WAM (upper row) with
    \sysname{} (lower row), with the sequence of frames progressing from left to right.}
    \label{fig:real_robot_sequences}
\end{figure}

\clearpage
\subsection{Additional Qualitative Examples}
\label{sec:appendix_qualitative}

Figures~\ref{fig:appendix_routing_group_a}--\ref{fig:appendix_routing_group_c}
extend the action-conditioned visual routing analysis in
Sec.~\ref{sec:qualitative_results} with six additional task examples.
Each task shows a reference setting on the left and six perturbations on the
right: background texture, lighting, robot initial state, camera viewpoint,
object layout, and sensor noise. Fast-WAM averages attention over Video-DiT
layers 0--29; \sysname{} uses the action-weighted LBQ routing map from the
final Video-DiT layer, following the procedure in the main text.

\begingroup
\setlength{\intextsep}{7pt}
\begin{figure}[H]
    \centering
    \includegraphics[width=\linewidth]{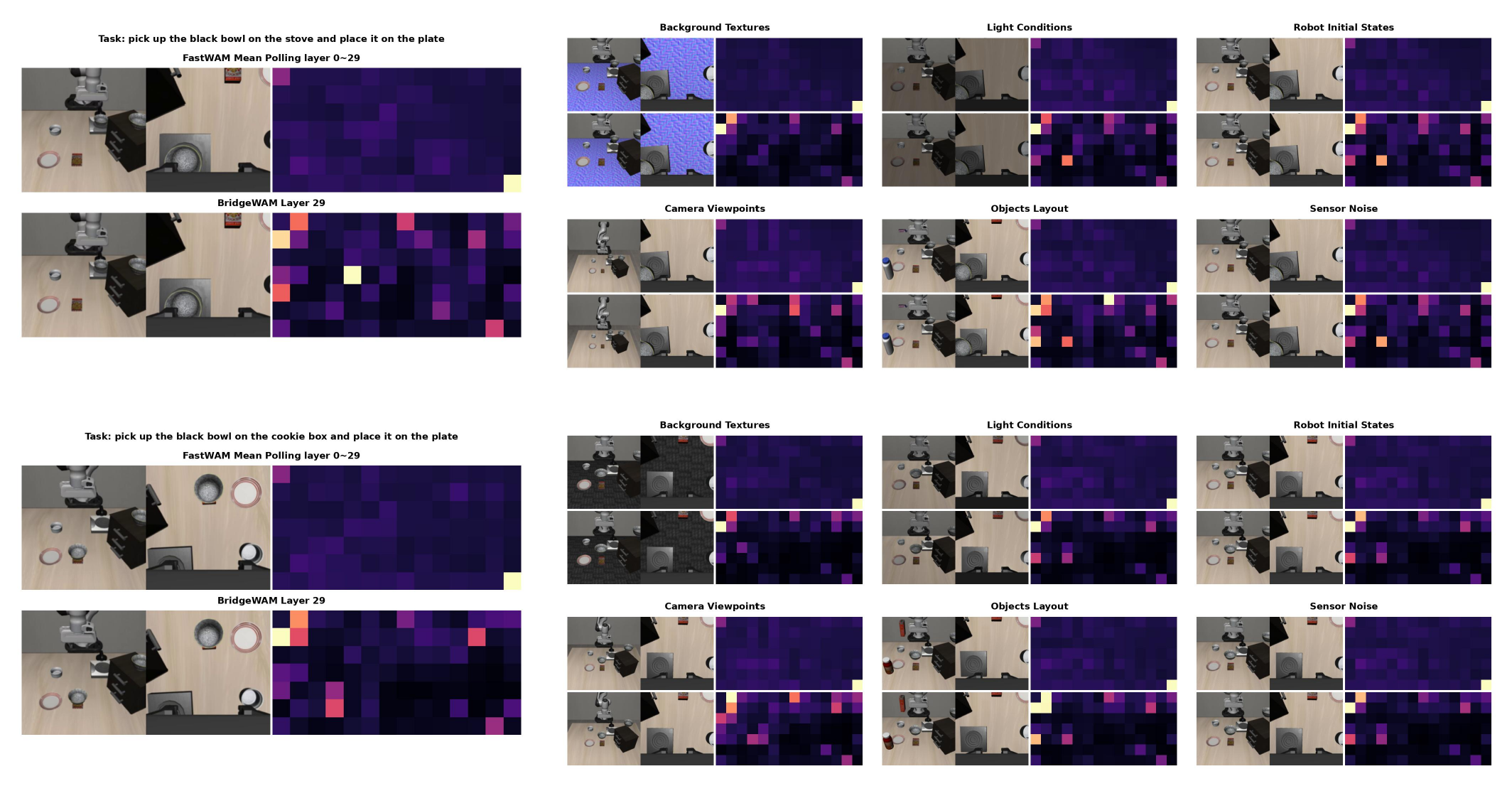}
    \caption{\textbf{Additional visual routing examples: bowl placement.}
    The tasks place a black bowl onto a plate from the stove (upper panel)
    or the cookie box (lower panel). Within each comparison, the results from Fast-WAM are shown above the corresponding results from \sysname{}.}
    \label{fig:appendix_routing_group_a}
\end{figure}

\begin{figure}[H]
    \centering
    \includegraphics[width=\linewidth]{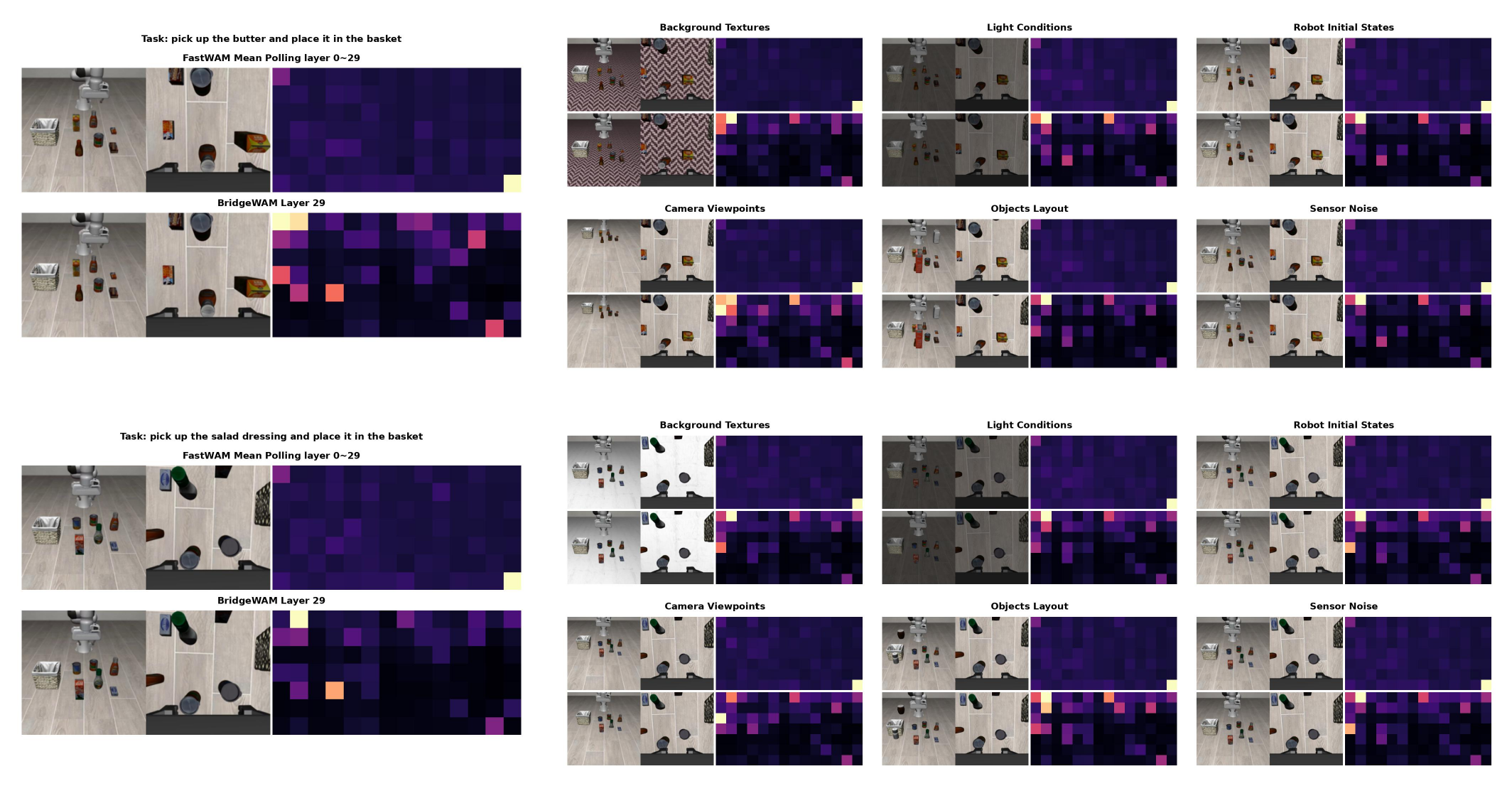}
    \caption{\textbf{Additional visual routing examples: basket placement.}
    The tasks place butter (upper panel) or salad dressing (lower panel)
    into a basket. Within each comparison, the results from Fast-WAM are shown above the corresponding results from \sysname{}.\looseness=-1}
    \label{fig:appendix_routing_group_b}
\end{figure}

\clearpage
\begin{figure}[H]
    \centering
    \includegraphics[width=\linewidth]{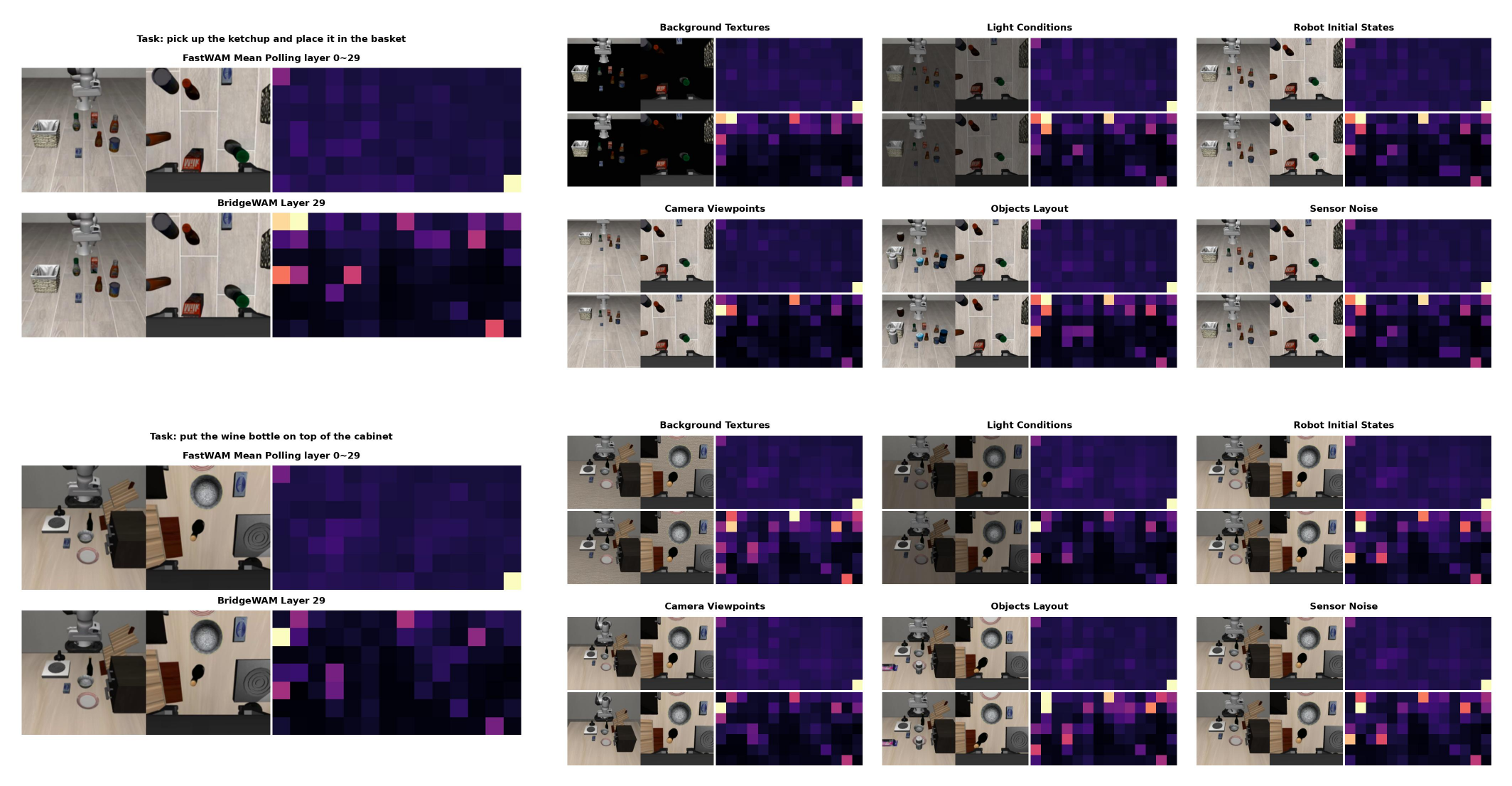}
    \caption{\textbf{Additional visual routing examples: different placement goals.}
    The tasks place ketchup into a basket (upper panel) and a wine bottle
    on top of a cabinet (lower panel). In each comparison, the Fast-WAM results are shown above the results for \sysname{}.}
    \label{fig:appendix_routing_group_c}
\end{figure}
\endgroup

Across these examples, \sysname{} exhibits more spatially differentiated
routing patterns than Fast-WAM, with the distribution of elevated weights
varying across tasks and perturbations. These comparisons complement
Fig.~\ref{fig:action_conditioned_routing} by examining visual readout across
different target objects and placement goals. As in the main analysis, the
maps visualize routing preferences; they neither establish the causal
contribution of individual visual features nor provide measurements of task success.\looseness=-1

\clearpage
\subsection{Complete Ablation Results}
\label{sec:appendix_ablation_results}

This section expands Tables~\ref{tab:action_head_depth}
and~\ref{tab:interface_ablation} with complete results for \sysname{} and
the ablation variants. Tables~\ref{tab:ablation_full_libero_spatial}--\ref{tab:ablation_full_libero_10}
report results for all 40 LIBERO tasks. Each configuration is evaluated for 50 episodes on each individual task, giving 500 episodes per suite and 2,000 episodes overall. The task indices used in these tables follow the zero-based task order within each benchmark suite.\looseness=-1

Tables~\ref{tab:ablation_plus_suite_depth} and~\ref{tab:ablation_plus_suite_interface}
break down LIBERO-Plus results by suite and perturbation category;
Tables~\ref{tab:ablation_plus_difficulty_depth}
and~\ref{tab:ablation_plus_difficulty_interface} further report the recorded
difficulty levels. All configurations complete the same 10,030 task variants
with one episode each: 2,402 Spatial, 2,518 Object, 2,591 Goal, and 2,519 Long
episodes. Pooled rates divide the total successful episodes by the total
episodes in the relevant group, rather than averaging rounded percentages.
The 121 lighting episodes without difficulty annotations are retained in all applicable totals and are also listed separately as Unspec.\looseness=-1

IDM ($x_0$ prefill) is the variant reported in
Table~\ref{tab:interface_ablation}. We additionally include the recorded
IDM (last video step) evaluation, which uses the final denoising forward
pass's LBQs without an extra $x_0$ prefill, as defined in
Appendix~\ref{sec:appendix_ablation}. The two IDM evaluations use the same
checkpoint; the results from these two evaluations are reported separately throughout these tables.

\begin{table}[H]
\centering
\caption{\textbf{Complete LIBERO-Spatial ablation results.} Success rates (\%) for all ten tasks, with 50 episodes per task. Task IDs follow the zero-based benchmark order; Avg. pools all 500 episodes.}
\label{tab:ablation_full_libero_spatial}
\footnotesize
\setlength{\tabcolsep}{3pt}
\renewcommand{\arraystretch}{1.03}
\begin{tabular}{@{}lrrrrrrrrrrr@{}}
\toprule
Method & 0 & 1 & 2 & 3 & 4 & 5 & 6 & 7 & 8 & 9 & Avg. \\
\midrule
\rowcolor{bridgewamrow!60!white}
\sysname{} (2 layers) & 100 & 100 & 100 & 100 & 98 & 94 & 100 & 100 & 98 & 100 & 99.00 \\
4 layers & 100 & 100 & 100 & 100 & 92 & 84 & 98 & 100 & 98 & 100 & 97.20 \\
8 layers & 100 & 98 & 100 & 98 & 96 & 86 & 100 & 100 & 96 & 100 & 97.40 \\
16 layers & 100 & 100 & 100 & 100 & 92 & 98 & 100 & 100 & 98 & 100 & 98.80 \\
30 layers (self-cross) & 100 & 94 & 100 & 100 & 98 & 92 & 100 & 100 & 94 & 100 & 97.80 \\
\midrule
Frozen-Video-DiT & 98 & 92 & 98 & 100 & 54 & 80 & 98 & 82 & 86 & 90 & 87.80 \\
MoT-readout & 100 & 98 & 100 & 98 & 98 & 90 & 100 & 100 & 96 & 100 & 98.00 \\
Text-state-LBQ-input & 98 & 96 & 100 & 100 & 92 & 90 & 96 & 100 & 94 & 98 & 96.40 \\
Joint & 100 & 100 & 100 & 98 & 92 & 88 & 100 & 96 & 100 & 100 & 97.40 \\
IDM ($x_0$ prefill) & 100 & 100 & 100 & 100 & 100 & 92 & 100 & 100 & 96 & 100 & 98.80 \\
\midrule
IDM (last video step) & 100 & 100 & 100 & 100 & 100 & 92 & 100 & 100 & 98 & 100 & 99.00 \\
\bottomrule
\end{tabular}
\end{table}
\begin{table}[H]
\centering
\caption{\textbf{Complete LIBERO-Object ablation results.} Success rates (\%) for all ten tasks, with 50 episodes per task. Task IDs follow the zero-based benchmark order; Avg. pools all 500 episodes.}
\label{tab:ablation_full_libero_object}
\footnotesize
\setlength{\tabcolsep}{3pt}
\renewcommand{\arraystretch}{1.03}
\begin{tabular}{@{}lrrrrrrrrrrr@{}}
\toprule
Method & 0 & 1 & 2 & 3 & 4 & 5 & 6 & 7 & 8 & 9 & Avg. \\
\midrule
\rowcolor{bridgewamrow!60!white}
\sysname{} (2 layers) & 100 & 100 & 100 & 100 & 100 & 100 & 100 & 100 & 100 & 100 & 100.00 \\
4 layers & 100 & 100 & 100 & 100 & 100 & 100 & 100 & 100 & 100 & 100 & 100.00 \\
8 layers & 100 & 100 & 100 & 100 & 100 & 100 & 100 & 100 & 98 & 100 & 99.80 \\
16 layers & 100 & 100 & 100 & 100 & 96 & 98 & 100 & 100 & 100 & 100 & 99.40 \\
30 layers (self-cross) & 100 & 100 & 100 & 100 & 100 & 100 & 100 & 100 & 100 & 100 & 100.00 \\
\midrule
Frozen-Video-DiT & 98 & 100 & 100 & 94 & 96 & 96 & 98 & 90 & 98 & 100 & 97.00 \\
MoT-readout & 100 & 100 & 100 & 98 & 100 & 98 & 100 & 100 & 100 & 98 & 99.40 \\
Text-state-LBQ-input & 100 & 100 & 100 & 100 & 98 & 98 & 100 & 100 & 96 & 100 & 99.20 \\
Joint & 100 & 100 & 100 & 100 & 98 & 100 & 100 & 100 & 100 & 98 & 99.60 \\
IDM ($x_0$ prefill) & 98 & 98 & 100 & 98 & 100 & 98 & 100 & 100 & 100 & 86 & 97.80 \\
\midrule
IDM (last video step) & 100 & 98 & 100 & 100 & 100 & 100 & 100 & 100 & 98 & 84 & 98.00 \\
\bottomrule
\end{tabular}
\end{table}
\clearpage
\begin{table}[H]
\centering
\caption{\textbf{Complete LIBERO-Goal ablation results.} Success rates (\%) for all ten tasks, with 50 episodes per task. Task IDs follow the zero-based benchmark order; Avg. pools all 500 episodes.}
\label{tab:ablation_full_libero_goal}
\footnotesize
\setlength{\tabcolsep}{3pt}
\renewcommand{\arraystretch}{1.03}
\begin{tabular}{@{}lrrrrrrrrrrr@{}}
\toprule
Method & 0 & 1 & 2 & 3 & 4 & 5 & 6 & 7 & 8 & 9 & Avg. \\
\midrule
\rowcolor{bridgewamrow!60!white}
\sysname{} (2 layers) & 98 & 98 & 88 & 96 & 100 & 96 & 100 & 98 & 100 & 100 & 97.40 \\
4 layers & 100 & 100 & 88 & 88 & 100 & 100 & 98 & 100 & 98 & 100 & 97.20 \\
8 layers & 98 & 100 & 88 & 100 & 98 & 100 & 100 & 98 & 100 & 94 & 97.60 \\
16 layers & 98 & 100 & 90 & 92 & 100 & 94 & 98 & 98 & 100 & 92 & 96.20 \\
30 layers (self-cross) & 100 & 100 & 94 & 94 & 96 & 98 & 100 & 100 & 100 & 94 & 97.60 \\
\midrule
Frozen-Video-DiT & 82 & 96 & 96 & 72 & 98 & 96 & 96 & 100 & 98 & 14 & 84.80 \\
MoT-readout & 98 & 100 & 94 & 100 & 96 & 100 & 98 & 100 & 98 & 90 & 97.40 \\
Text-state-LBQ-input & 100 & 100 & 94 & 98 & 96 & 100 & 100 & 100 & 100 & 82 & 97.00 \\
Joint & 100 & 100 & 84 & 100 & 96 & 100 & 100 & 100 & 100 & 86 & 96.60 \\
IDM ($x_0$ prefill) & 100 & 100 & 96 & 100 & 100 & 100 & 96 & 100 & 100 & 98 & 99.00 \\
\midrule
IDM (last video step) & 98 & 100 & 86 & 96 & 98 & 98 & 100 & 100 & 100 & 90 & 96.60 \\
\bottomrule
\end{tabular}
\end{table}
\begin{table}[H]
\centering
\caption{\textbf{Complete LIBERO-Long ablation results.} Success rates (\%) for all ten tasks, with 50 episodes per task. Task IDs follow the zero-based benchmark order; Avg. pools all 500 episodes.}
\label{tab:ablation_full_libero_10}
\footnotesize
\setlength{\tabcolsep}{3pt}
\renewcommand{\arraystretch}{1.03}
\begin{tabular}{@{}lrrrrrrrrrrr@{}}
\toprule
Method & 0 & 1 & 2 & 3 & 4 & 5 & 6 & 7 & 8 & 9 & Avg. \\
\midrule
\rowcolor{bridgewamrow!60!white}
\sysname{} (2 layers) & 98 & 100 & 90 & 100 & 100 & 100 & 86 & 96 & 96 & 100 & 96.60 \\
4 layers & 96 & 100 & 98 & 100 & 90 & 98 & 90 & 96 & 98 & 100 & 96.60 \\
8 layers & 98 & 100 & 92 & 100 & 92 & 100 & 92 & 96 & 100 & 98 & 96.80 \\
16 layers & 96 & 98 & 96 & 100 & 92 & 100 & 82 & 90 & 84 & 96 & 93.40 \\
30 layers (self-cross) & 96 & 98 & 96 & 100 & 86 & 100 & 86 & 98 & 88 & 100 & 94.80 \\
\midrule
Frozen-Video-DiT & 74 & 74 & 96 & 96 & 32 & 90 & 38 & 66 & 88 & 70 & 72.40 \\
MoT-readout & 94 & 100 & 94 & 100 & 94 & 98 & 86 & 100 & 98 & 100 & 96.40 \\
Text-state-LBQ-input & 98 & 100 & 92 & 100 & 90 & 100 & 92 & 98 & 94 & 98 & 96.20 \\
Joint & 96 & 100 & 100 & 98 & 96 & 100 & 90 & 100 & 94 & 94 & 96.80 \\
IDM ($x_0$ prefill) & 96 & 100 & 94 & 100 & 94 & 100 & 90 & 100 & 94 & 98 & 96.60 \\
\midrule
IDM (last video step) & 100 & 100 & 96 & 98 & 96 & 100 & 88 & 100 & 90 & 98 & 96.60 \\
\bottomrule
\end{tabular}
\end{table}
\clearpage
\begin{table}[H]
\centering
\caption{\textbf{Complete LIBERO-Plus action-head depth results by suite and perturbation.} The entries in this table are task success rates (\%); pooled rates use episode counts, not averages of the displayed percentages. N denotes the total number of episodes pooled across all suites.}
\label{tab:ablation_plus_suite_depth}
\footnotesize
\setlength{\tabcolsep}{3pt}
\renewcommand{\arraystretch}{1.03}
\begin{tabular}{@{}lrrrrrr@{}}
\toprule
Perturbation & Spatial & Object & Goal & Long & Pooled & N \\
\midrule
\rowcolor{bridgewamrow!60!white}
\multicolumn{7}{@{}l}{\textbf{\sysname{} (2 layers)}} \\
Camera viewpoint & 65.16 & 83.59 & 30.64 & 46.78 & 56.10 & 1,599 \\
Robot initial state & 60.29 & 62.81 & 51.83 & 66.92 & 60.39 & 1,550 \\
Language instruction & 86.15 & 95.76 & 69.76 & 88.77 & 84.65 & 1,537 \\
Lighting & 97.26 & 95.29 & 94.98 & 81.39 & 92.38 & 1,142 \\
Background texture & 84.50 & 86.29 & 64.41 & 41.52 & 68.12 & 1,076 \\
Sensor noise & 76.35 & 91.94 & 60.69 & 65.26 & 73.64 & 1,601 \\
Object layout & 78.70 & 79.90 & 61.88 & 82.37 & 75.08 & 1,525 \\
\textit{All perturbations} & 77.64 & 84.47 & 60.29 & 67.17 & 72.24 & 10,030 \\
\midrule
\multicolumn{7}{@{}l}{\textbf{4 layers}} \\
Camera viewpoint & 35.64 & 83.84 & 24.26 & 27.68 & 42.59 & 1,599 \\
Robot initial state & 52.86 & 64.57 & 39.61 & 58.78 & 53.87 & 1,550 \\
Language instruction & 83.33 & 92.94 & 67.07 & 82.51 & 81.00 & 1,537 \\
Lighting & 97.60 & 94.28 & 97.13 & 84.31 & 93.43 & 1,142 \\
Background texture & 75.19 & 89.11 & 54.80 & 37.02 & 62.83 & 1,076 \\
Sensor noise & 62.68 & 92.42 & 50.40 & 62.58 & 67.58 & 1,601 \\
Object layout & 71.95 & 79.90 & 60.00 & 75.64 & 71.48 & 1,525 \\
\textit{All perturbations} & 67.44 & 84.63 & 54.30 & 60.26 & 66.56 & 10,030 \\
\midrule
\multicolumn{7}{@{}l}{\textbf{8 layers}} \\
Camera viewpoint & 47.34 & 77.02 & 20.59 & 28.88 & 43.03 & 1,599 \\
Robot initial state & 50.00 & 61.31 & 35.94 & 57.00 & 50.97 & 1,550 \\
Language instruction & 71.79 & 98.59 & 47.07 & 73.89 & 71.89 & 1,537 \\
Lighting & 97.26 & 98.65 & 92.47 & 69.34 & 89.75 & 1,142 \\
Background texture & 81.40 & 87.10 & 54.80 & 43.60 & 65.61 & 1,076 \\
Sensor noise & 73.79 & 91.47 & 61.21 & 65.48 & 73.14 & 1,601 \\
Object layout & 81.04 & 78.41 & 52.47 & 75.00 & 71.15 & 1,525 \\
\textit{All perturbations} & 70.69 & 83.76 & 49.83 & 58.44 & 65.50 & 10,030 \\
\midrule
\multicolumn{7}{@{}l}{\textbf{16 layers}} \\
Camera viewpoint & 37.50 & 66.41 & 34.07 & 24.34 & 40.34 & 1,599 \\
Robot initial state & 46.86 & 59.30 & 39.12 & 61.58 & 51.74 & 1,550 \\
Language instruction & 70.26 & 93.50 & 63.90 & 86.68 & 78.01 & 1,537 \\
Lighting & 95.89 & 98.65 & 73.12 & 84.67 & 88.35 & 1,142 \\
Background texture & 75.58 & 77.02 & 49.47 & 41.52 & 59.94 & 1,076 \\
Sensor noise & 80.06 & 95.26 & 62.27 & 63.25 & 75.14 & 1,601 \\
Object layout & 77.14 & 78.41 & 51.06 & 81.09 & 71.02 & 1,525 \\
\textit{All perturbations} & 67.94 & 80.70 & 52.37 & 62.13 & 65.66 & 10,030 \\
\midrule
\multicolumn{7}{@{}l}{\textbf{30 layers (self-cross)}} \\
Camera viewpoint & 58.24 & 61.36 & 30.39 & 40.33 & 47.22 & 1,599 \\
Robot initial state & 43.71 & 57.04 & 43.03 & 59.80 & 51.03 & 1,550 \\
Language instruction & 65.13 & 94.07 & 73.41 & 80.68 & 77.88 & 1,537 \\
Lighting & 89.38 & 97.98 & 97.49 & 90.15 & 93.78 & 1,142 \\
Background texture & 72.09 & 75.40 & 69.04 & 42.56 & 64.13 & 1,076 \\
Sensor noise & 82.91 & 87.44 & 65.44 & 65.92 & 75.20 & 1,601 \\
Object layout & 78.70 & 78.16 & 56.94 & 77.24 & 72.20 & 1,525 \\
\textit{All perturbations} & 69.40 & 78.04 & 60.09 & 64.31 & 67.89 & 10,030 \\
\bottomrule
\end{tabular}
\end{table}
\clearpage
\begin{table}[H]
\centering
\caption{\textbf{Complete LIBERO-Plus interface and coupling results by suite and perturbation.} Entries are success rates (\%); pooled rates use episode counts, not averages of the displayed percentages. N denotes the total number of episodes across all the evaluated suites.\looseness=-1}
\label{tab:ablation_plus_suite_interface}
\footnotesize
\setlength{\tabcolsep}{3pt}
\renewcommand{\arraystretch}{1.03}
\begin{tabular}{@{}lrrrrrr@{}}
\toprule
Perturbation & Spatial & Object & Goal & Long & Pooled & N \\
\midrule
\multicolumn{7}{@{}l}{\textbf{Frozen-Video-DiT}} \\
Camera viewpoint & 3.72 & 0.51 & 2.21 & 0.24 & 1.63 & 1,599 \\
Robot initial state & 32.00 & 33.92 & 8.56 & 20.36 & 23.35 & 1,550 \\
Language instruction & 56.41 & 82.77 & 39.51 & 50.65 & 56.54 & 1,537 \\
Lighting & 53.08 & 87.88 & 22.94 & 15.69 & 45.80 & 1,142 \\
Background texture & 20.16 & 6.05 & 2.85 & 0.00 & 6.97 & 1,076 \\
Sensor noise & 6.84 & 9.48 & 5.01 & 4.01 & 6.31 & 1,601 \\
Object layout & 51.17 & 55.58 & 24.24 & 24.36 & 39.34 & 1,525 \\
\textit{All perturbations} & 32.22 & 38.52 & 15.44 & 16.36 & 25.48 & 10,030 \\
\midrule
\multicolumn{7}{@{}l}{\textbf{MoT-readout}} \\
Camera viewpoint & 42.29 & 61.62 & 10.78 & 21.48 & 33.58 & 1,599 \\
Robot initial state & 54.57 & 66.33 & 30.32 & 52.67 & 50.71 & 1,550 \\
Language instruction & 80.26 & 94.07 & 41.71 & 74.93 & 71.83 & 1,537 \\
Lighting & 98.63 & 95.29 & 79.57 & 70.44 & 86.34 & 1,142 \\
Background texture & 89.53 & 87.50 & 43.42 & 45.33 & 65.15 & 1,076 \\
Sensor noise & 70.09 & 89.57 & 29.02 & 54.57 & 61.15 & 1,601 \\
Object layout & 75.58 & 77.67 & 33.41 & 72.76 & 63.80 & 1,525 \\
\textit{All perturbations} & 71.57 & 80.70 & 36.09 & 54.78 & 60.48 & 10,030 \\
\midrule
\multicolumn{7}{@{}l}{\textbf{Text-state-LBQ-input}} \\
Camera viewpoint & 48.67 & 72.73 & 24.75 & 19.09 & 40.78 & 1,599 \\
Robot initial state & 48.86 & 59.30 & 26.41 & 50.64 & 46.06 & 1,550 \\
Language instruction & 64.87 & 92.37 & 48.05 & 70.23 & 68.05 & 1,537 \\
Lighting & 91.78 & 96.30 & 87.81 & 67.15 & 86.08 & 1,142 \\
Background texture & 71.71 & 77.42 & 43.77 & 33.56 & 55.48 & 1,076 \\
Sensor noise & 62.39 & 85.78 & 40.63 & 47.22 & 59.15 & 1,601 \\
Object layout & 71.43 & 80.15 & 50.59 & 62.50 & 66.10 & 1,525 \\
\textit{All perturbations} & 64.70 & 79.98 & 44.11 & 49.07 & 59.29 & 10,030 \\
\midrule
\multicolumn{7}{@{}l}{\textbf{Joint}} \\
Camera viewpoint & 52.93 & 50.76 & 29.90 & 19.09 & 37.65 & 1,599 \\
Robot initial state & 65.71 & 57.04 & 31.30 & 66.16 & 54.52 & 1,550 \\
Language instruction & 90.26 & 98.02 & 63.90 & 91.38 & 85.30 & 1,537 \\
Lighting & 96.92 & 97.31 & 77.06 & 80.66 & 88.27 & 1,142 \\
Background texture & 71.32 & 61.69 & 33.45 & 31.14 & 48.42 & 1,076 \\
Sensor noise & 71.23 & 66.59 & 35.88 & 49.44 & 55.53 & 1,601 \\
Object layout & 88.31 & 78.91 & 58.35 & 89.10 & 77.64 & 1,525 \\
\textit{All perturbations} & 76.52 & 72.12 & 46.51 & 59.59 & 63.41 & 10,030 \\
\midrule
\multicolumn{7}{@{}l}{\textbf{IDM ($x_0$ prefill)}} \\
Camera viewpoint & 52.93 & 54.29 & 34.56 & 31.74 & 43.03 & 1,599 \\
Robot initial state & 67.71 & 51.51 & 52.32 & 71.25 & 60.39 & 1,550 \\
Language instruction & 93.59 & 99.44 & 84.63 & 98.69 & 93.82 & 1,537 \\
Lighting & 95.21 & 88.89 & 93.55 & 73.72 & 88.00 & 1,142 \\
Background texture & 72.09 & 66.13 & 48.04 & 33.91 & 54.18 & 1,076 \\
Sensor noise & 65.24 & 65.64 & 46.97 & 55.46 & 58.28 & 1,601 \\
Object layout & 87.53 & 80.15 & 56.71 & 88.78 & 77.25 & 1,525 \\
\textit{All perturbations} & 76.23 & 71.49 & 58.55 & 64.19 & 67.45 & 10,030 \\
\midrule
\multicolumn{7}{@{}l}{\textbf{IDM (last video step)}} \\
Camera viewpoint & 52.93 & 55.81 & 30.88 & 28.16 & 41.53 & 1,599 \\
Robot initial state & 66.29 & 50.75 & 49.14 & 72.01 & 59.23 & 1,550 \\
Language instruction & 94.62 & 99.15 & 83.17 & 96.34 & 93.04 & 1,537 \\
Lighting & 98.63 & 87.88 & 90.68 & 71.17 & 87.30 & 1,142 \\
Background texture & 72.87 & 65.73 & 49.11 & 32.87 & 54.28 & 1,076 \\
Sensor noise & 59.83 & 67.77 & 49.08 & 55.46 & 58.15 & 1,601 \\
Object layout & 88.05 & 80.89 & 55.53 & 87.18 & 76.92 & 1,525 \\
\textit{All perturbations} & 75.98 & 71.88 & 57.16 & 62.76 & 66.77 & 10,030 \\
\bottomrule
\end{tabular}
\end{table}
\clearpage
\begin{table}[H]
\centering
\caption{\textbf{Complete LIBERO-Plus action-head depth results by difficulty.} Rates (\%) are pooled across all four suites. Unspec. includes 121 lighting episodes without a difficulty annotation; all of these episodes are retained in the pooled totals. -- denotes cells with no evaluated episodes.}
\label{tab:ablation_plus_difficulty_depth}
\footnotesize
\setlength{\tabcolsep}{3pt}
\renewcommand{\arraystretch}{1.03}
\begin{tabular}{@{}lrrrrrrr@{}}
\toprule
Perturbation & Level 1 & Level 2 & Level 3 & Level 4 & Level 5 & Unspec. & Pooled \\
\midrule
\rowcolor{bridgewamrow!60!white}
\multicolumn{8}{@{}l}{\textbf{\sysname{} (2 layers)}} \\
Camera viewpoint & 70.85 & 60.28 & 56.92 & 57.89 & 45.59 & -- & 56.10 \\
Robot initial state & 87.50 & 82.19 & 64.97 & 45.26 & 30.48 & -- & 60.39 \\
Language instruction & 90.85 & 93.57 & 92.28 & 85.67 & 64.62 & -- & 84.65 \\
Lighting & 98.20 & 94.42 & 94.67 & 91.74 & 83.26 & 97.52 & 92.38 \\
Background texture & 87.55 & 78.23 & 69.58 & 55.21 & 30.14 & -- & 68.12 \\
Sensor noise & 81.36 & 84.60 & 80.00 & 70.50 & 42.32 & -- & 73.64 \\
Object layout & 86.49 & 86.75 & 78.42 & 71.60 & 48.44 & -- & 75.08 \\
\textit{All perturbations} & 86.01 & 81.83 & 75.41 & 69.35 & 49.21 & 97.52 & 72.24 \\
\midrule
\multicolumn{8}{@{}l}{\textbf{4 layers}} \\
Camera viewpoint & 58.29 & 50.83 & 44.62 & 33.33 & 33.06 & -- & 42.59 \\
Robot initial state & 80.65 & 70.55 & 59.89 & 42.46 & 23.93 & -- & 53.87 \\
Language instruction & 90.24 & 85.71 & 89.96 & 78.05 & 64.33 & -- & 81.00 \\
Lighting & 94.59 & 97.46 & 94.26 & 90.91 & 88.11 & 99.17 & 93.43 \\
Background texture & 78.54 & 73.43 & 64.64 & 47.24 & 32.19 & -- & 62.83 \\
Sensor noise & 69.49 & 78.80 & 73.67 & 63.45 & 41.91 & -- & 67.58 \\
Object layout & 83.62 & 83.11 & 71.43 & 63.81 & 51.56 & -- & 71.48 \\
\textit{All perturbations} & 79.93 & 75.61 & 69.63 & 61.35 & 46.18 & 99.17 & 66.56 \\
\midrule
\multicolumn{8}{@{}l}{\textbf{8 layers}} \\
Camera viewpoint & 63.32 & 49.17 & 41.85 & 39.47 & 32.65 & -- & 43.03 \\
Robot initial state & 80.65 & 70.21 & 54.81 & 38.60 & 19.94 & -- & 50.97 \\
Language instruction & 82.62 & 77.14 & 79.15 & 71.04 & 52.63 & -- & 71.89 \\
Lighting & 93.69 & 89.85 & 91.39 & 90.91 & 83.26 & 92.56 & 89.75 \\
Background texture & 82.40 & 70.48 & 66.54 & 53.99 & 41.10 & -- & 65.61 \\
Sensor noise & 72.32 & 79.80 & 76.67 & 73.89 & 54.36 & -- & 73.14 \\
Object layout & 78.74 & 82.78 & 77.20 & 64.59 & 48.79 & -- & 71.15 \\
\textit{All perturbations} & 78.77 & 73.34 & 68.19 & 63.10 & 44.65 & 92.56 & 65.50 \\
\midrule
\multicolumn{8}{@{}l}{\textbf{16 layers}} \\
Camera viewpoint & 59.80 & 51.94 & 46.46 & 40.35 & 19.71 & -- & 40.34 \\
Robot initial state & 85.08 & 71.92 & 56.15 & 33.68 & 21.37 & -- & 51.74 \\
Language instruction & 84.15 & 81.43 & 84.94 & 79.27 & 62.87 & -- & 78.01 \\
Lighting & 89.19 & 91.88 & 93.03 & 90.08 & 83.26 & 78.51 & 88.35 \\
Background texture & 78.97 & 71.59 & 56.65 & 43.56 & 32.19 & -- & 59.94 \\
Sensor noise & 81.92 & 85.00 & 82.67 & 72.58 & 44.40 & -- & 75.14 \\
Object layout & 84.20 & 78.15 & 72.04 & 64.98 & 51.90 & -- & 71.02 \\
\textit{All perturbations} & 80.72 & 75.43 & 68.86 & 62.67 & 42.20 & 78.51 & 65.66 \\
\midrule
\multicolumn{8}{@{}l}{\textbf{30 layers (self-cross)}} \\
Camera viewpoint & 66.83 & 57.50 & 54.15 & 42.98 & 28.95 & -- & 47.22 \\
Robot initial state & 77.02 & 70.89 & 56.15 & 38.60 & 20.80 & -- & 51.03 \\
Language instruction & 82.01 & 80.36 & 84.17 & 77.13 & 67.84 & -- & 77.88 \\
Lighting & 98.20 & 95.43 & 91.80 & 92.56 & 91.63 & 97.52 & 93.78 \\
Background texture & 82.83 & 70.11 & 62.36 & 52.15 & 39.73 & -- & 64.13 \\
Sensor noise & 78.53 & 82.40 & 83.33 & 71.28 & 53.94 & -- & 75.20 \\
Object layout & 83.62 & 83.77 & 74.47 & 64.20 & 50.87 & -- & 72.20 \\
\textit{All perturbations} & 80.60 & 76.39 & 71.01 & 64.05 & 47.48 & 97.52 & 67.89 \\
\bottomrule
\end{tabular}
\end{table}
\clearpage
\begin{table}[H]
\centering
\caption{\textbf{Complete LIBERO-Plus interface and coupling results by difficulty.} Rates (\%) are computed by pooling episodes across all four suites. Unspec. includes 121 lighting episodes without a difficulty annotation; these episodes are retained in pooled totals. -- denotes no episodes.\looseness=-1}
\label{tab:ablation_plus_difficulty_interface}
\footnotesize
\setlength{\tabcolsep}{3pt}
\renewcommand{\arraystretch}{1.03}
\begin{tabular}{@{}lrrrrrrr@{}}
\toprule
Perturbation & Level 1 & Level 2 & Level 3 & Level 4 & Level 5 & Unspec. & Pooled \\
\midrule
\multicolumn{8}{@{}l}{\textbf{Frozen-Video-DiT}} \\
Camera viewpoint & 6.03 & 2.78 & 0.62 & 0.44 & 0.21 & -- & 1.63 \\
Robot initial state & 41.94 & 34.25 & 24.33 & 12.28 & 9.12 & -- & 23.35 \\
Language instruction & 64.02 & 62.50 & 59.85 & 54.27 & 44.15 & -- & 56.54 \\
Lighting & 47.75 & 44.16 & 56.56 & 50.83 & 41.41 & 23.14 & 45.80 \\
Background texture & 14.16 & 7.75 & 6.46 & 2.45 & 0.00 & -- & 6.97 \\
Sensor noise & 16.95 & 10.80 & 3.33 & 1.83 & 0.00 & -- & 6.31 \\
Object layout & 45.11 & 43.38 & 40.43 & 39.69 & 26.64 & -- & 39.34 \\
\textit{All perturbations} & 36.44 & 26.25 & 26.07 & 23.86 & 17.04 & 23.14 & 25.48 \\
\midrule
\multicolumn{8}{@{}l}{\textbf{MoT-readout}} \\
Camera viewpoint & 48.24 & 36.67 & 32.00 & 30.26 & 27.93 & -- & 33.58 \\
Robot initial state & 79.84 & 71.23 & 52.41 & 36.49 & 22.79 & -- & 50.71 \\
Language instruction & 84.15 & 76.43 & 77.22 & 72.87 & 51.17 & -- & 71.83 \\
Lighting & 84.68 & 91.37 & 90.16 & 85.95 & 78.85 & 86.78 & 86.34 \\
Background texture & 82.83 & 71.59 & 65.02 & 52.15 & 39.73 & -- & 65.15 \\
Sensor noise & 64.97 & 72.00 & 65.33 & 60.57 & 31.54 & -- & 61.15 \\
Object layout & 67.53 & 70.86 & 69.60 & 64.59 & 44.64 & -- & 63.80 \\
\textit{All perturbations} & 73.42 & 68.21 & 62.85 & 58.48 & 39.99 & 86.78 & 60.48 \\
\midrule
\multicolumn{8}{@{}l}{\textbf{Text-state-LBQ-input}} \\
Camera viewpoint & 61.81 & 52.22 & 42.15 & 32.46 & 26.69 & -- & 40.78 \\
Robot initial state & 68.95 & 62.33 & 49.47 & 34.74 & 21.94 & -- & 46.06 \\
Language instruction & 76.52 & 71.43 & 75.68 & 70.43 & 49.12 & -- & 68.05 \\
Lighting & 92.79 & 90.86 & 86.48 & 89.67 & 73.57 & 87.60 & 86.08 \\
Background texture & 73.39 & 61.99 & 53.99 & 47.24 & 26.71 & -- & 55.48 \\
Sensor noise & 70.62 & 70.60 & 62.33 & 56.14 & 27.80 & -- & 59.15 \\
Object layout & 72.99 & 72.19 & 69.00 & 63.42 & 50.52 & -- & 66.10 \\
\textit{All perturbations} & 72.87 & 67.57 & 61.37 & 57.05 & 38.12 & 87.60 & 59.29 \\
\midrule
\multicolumn{8}{@{}l}{\textbf{Joint}} \\
Camera viewpoint & 74.87 & 57.78 & 34.15 & 27.63 & 14.58 & -- & 37.65 \\
Robot initial state & 88.31 & 75.34 & 60.70 & 37.19 & 20.80 & -- & 54.52 \\
Language instruction & 91.77 & 91.07 & 93.44 & 83.54 & 69.88 & -- & 85.30 \\
Lighting & 90.09 & 88.32 & 95.08 & 90.08 & 82.38 & 80.17 & 88.27 \\
Background texture & 72.96 & 60.15 & 44.49 & 31.29 & 13.70 & -- & 48.42 \\
Sensor noise & 79.66 & 74.20 & 62.00 & 39.16 & 17.01 & -- & 55.53 \\
Object layout & 89.94 & 90.73 & 85.41 & 67.70 & 49.13 & -- & 77.64 \\
\textit{All perturbations} & 84.73 & 75.61 & 66.67 & 54.93 & 37.11 & 80.17 & 63.41 \\
\midrule
\multicolumn{8}{@{}l}{\textbf{IDM ($x_0$ prefill)}} \\
Camera viewpoint & 74.37 & 58.61 & 46.46 & 34.21 & 20.53 & -- & 43.03 \\
Robot initial state & 93.15 & 80.82 & 64.97 & 44.21 & 28.49 & -- & 60.39 \\
Language instruction & 96.95 & 97.50 & 95.75 & 94.21 & 85.96 & -- & 93.82 \\
Lighting & 96.40 & 92.89 & 90.57 & 83.47 & 77.53 & 95.87 & 88.00 \\
Background texture & 74.25 & 66.42 & 49.05 & 42.33 & 21.92 & -- & 54.18 \\
Sensor noise & 83.62 & 80.00 & 63.33 & 37.08 & 21.99 & -- & 58.28 \\
Object layout & 89.37 & 85.76 & 80.85 & 75.10 & 51.56 & -- & 77.25 \\
\textit{All perturbations} & 87.35 & 79.11 & 69.15 & 59.33 & 43.40 & 95.87 & 67.45 \\
\midrule
\multicolumn{8}{@{}l}{\textbf{IDM (last video step)}} \\
Camera viewpoint & 75.88 & 60.56 & 41.85 & 31.14 & 18.07 & -- & 41.53 \\
Robot initial state & 90.32 & 80.14 & 63.64 & 42.11 & 29.06 & -- & 59.23 \\
Language instruction & 97.26 & 97.50 & 93.82 & 94.21 & 83.63 & -- & 93.04 \\
Lighting & 96.40 & 90.86 & 93.44 & 79.34 & 77.09 & 95.87 & 87.30 \\
Background texture & 76.82 & 66.05 & 49.81 & 39.88 & 20.55 & -- & 54.28 \\
Sensor noise & 85.31 & 79.60 & 59.33 & 39.43 & 21.99 & -- & 58.15 \\
Object layout & 87.93 & 84.44 & 80.55 & 76.65 & 51.90 & -- & 76.92 \\
\textit{All perturbations} & 87.41 & 78.84 & 67.77 & 58.59 & 42.44 & 95.87 & 66.77 \\
\bottomrule
\end{tabular}
\end{table}

\clearpage
\paragraph{LBQ sequence-length details.}
Tables~\ref{tab:lbq_full_libero}--\ref{tab:lbq_plus_difficulty_2}
expand the 1-, 8-, 16-, 24-, 36-, and 64-LBQ configurations reported in
Table~\ref{tab:lbq_length_ablation}. The default 32-LBQ model and
Action Expert depth variants are detailed in the preceding ablation tables.
The additional tables report results for all LIBERO tasks and break down LIBERO-Plus
results by suite, perturbation category, and difficulty, using the same evaluation protocol and the same pooled aggregation procedure defined above.

\begin{table}[H]
\centering
\caption{\textbf{LBQ sequence-length ablations on all LIBERO tasks.} Detailed results for the length variants in Table~\ref{tab:lbq_length_ablation}. Entries are success rates (\%); each task has 50 episodes. Task IDs follow the zero-based ordering of tasks within each suite. Avg. pools 500 episodes per suite.}
\label{tab:lbq_full_libero}
\footnotesize
\setlength{\tabcolsep}{3pt}
\renewcommand{\arraystretch}{1.03}
\begin{tabular}{@{}lrrrrrrrrrrr@{}}
\toprule
LBQs & 0 & 1 & 2 & 3 & 4 & 5 & 6 & 7 & 8 & 9 & Avg. \\
\midrule
\multicolumn{12}{@{}l}{\textbf{LIBERO-Spatial}} \\
1 & 100 & 98 & 100 & 100 & 96 & 94 & 100 & 100 & 98 & 100 & 98.60 \\
8 & 100 & 96 & 100 & 100 & 96 & 92 & 98 & 100 & 98 & 100 & 98.00 \\
16 & 100 & 98 & 100 & 100 & 92 & 98 & 100 & 100 & 96 & 98 & 98.20 \\
24 & 98 & 100 & 100 & 98 & 100 & 88 & 100 & 100 & 98 & 100 & 98.20 \\
36 & 100 & 96 & 100 & 100 & 98 & 86 & 100 & 98 & 98 & 100 & 97.60 \\
64 & 100 & 96 & 100 & 100 & 96 & 98 & 100 & 98 & 96 & 100 & 98.40 \\
\midrule
\multicolumn{12}{@{}l}{\textbf{LIBERO-Object}} \\
1 & 100 & 100 & 100 & 98 & 100 & 98 & 100 & 100 & 100 & 100 & 99.60 \\
8 & 100 & 100 & 100 & 100 & 100 & 100 & 100 & 100 & 98 & 100 & 99.80 \\
16 & 100 & 100 & 100 & 100 & 100 & 96 & 100 & 100 & 98 & 100 & 99.40 \\
24 & 100 & 100 & 100 & 98 & 100 & 100 & 100 & 100 & 100 & 98 & 99.60 \\
36 & 100 & 100 & 100 & 100 & 100 & 100 & 100 & 100 & 98 & 100 & 99.80 \\
64 & 100 & 100 & 100 & 100 & 100 & 100 & 98 & 100 & 100 & 100 & 99.80 \\
\midrule
\multicolumn{12}{@{}l}{\textbf{LIBERO-Goal}} \\
1 & 96 & 98 & 86 & 96 & 98 & 98 & 98 & 100 & 96 & 88 & 95.40 \\
8 & 96 & 98 & 96 & 92 & 98 & 100 & 100 & 100 & 98 & 94 & 97.20 \\
16 & 100 & 98 & 88 & 100 & 100 & 100 & 100 & 100 & 100 & 98 & 98.40 \\
24 & 100 & 100 & 86 & 94 & 100 & 98 & 100 & 100 & 94 & 94 & 96.60 \\
36 & 100 & 98 & 94 & 90 & 100 & 98 & 100 & 100 & 96 & 98 & 97.40 \\
64 & 100 & 98 & 94 & 96 & 100 & 100 & 100 & 100 & 100 & 94 & 98.20 \\
\midrule
\multicolumn{12}{@{}l}{\textbf{LIBERO-Long}} \\
1 & 94 & 94 & 84 & 94 & 90 & 100 & 80 & 100 & 94 & 90 & 92.00 \\
8 & 92 & 100 & 96 & 96 & 92 & 100 & 92 & 98 & 98 & 98 & 96.20 \\
16 & 96 & 100 & 86 & 100 & 92 & 100 & 92 & 100 & 100 & 98 & 96.40 \\
24 & 96 & 98 & 88 & 100 & 96 & 100 & 84 & 100 & 96 & 98 & 95.60 \\
36 & 94 & 96 & 90 & 98 & 98 & 100 & 98 & 100 & 94 & 98 & 96.60 \\
64 & 98 & 100 & 92 & 98 & 86 & 98 & 86 & 98 & 94 & 98 & 94.80 \\
\bottomrule
\end{tabular}
\end{table}

\clearpage
\begin{table}[H]
\centering
\caption{\textbf{LIBERO-Plus LBQ sequence-length results by suite and perturbation (1/2).} Success rates (\%) for 1, 8, 16 LBQs. Pooled rates use episode counts; $N$ counts episodes across suites.}
\label{tab:lbq_plus_suite_1}
\footnotesize
\setlength{\tabcolsep}{3pt}
\renewcommand{\arraystretch}{1.03}
\begin{tabular}{@{}lrrrrrr@{}}
\toprule
Perturbation & Spatial & Object & Goal & Long & Pooled & $N$ \\
\midrule
\multicolumn{7}{@{}l}{\textbf{1 LBQ}} \\
Camera viewpoint & 35.37 & 76.26 & 17.16 & 26.25 & 38.46 & 1,599 \\
Robot initial state & 65.14 & 66.58 & 44.99 & 61.58 & 59.29 & 1,550 \\
Language instruction & 87.69 & 95.20 & 71.71 & 82.51 & 83.86 & 1,537 \\
Lighting & 92.12 & 90.57 & 88.53 & 68.98 & 85.29 & 1,142 \\
Background texture & 68.99 & 72.58 & 37.37 & 24.57 & 49.63 & 1,076 \\
Sensor noise & 71.79 & 91.47 & 48.28 & 65.03 & 69.52 & 1,601 \\
Object layout & 84.16 & 84.12 & 51.06 & 74.36 & 72.92 & 1,525 \\
\textit{All perturbations} & 71.86 & 82.53 & 50.17 & 57.64 & 65.36 & 10,030 \\
\midrule
\multicolumn{7}{@{}l}{\textbf{8 LBQs}} \\
Camera viewpoint & 50.80 & 69.19 & 31.37 & 25.78 & 43.84 & 1,599 \\
Robot initial state & 49.71 & 55.28 & 34.72 & 60.56 & 49.94 & 1,550 \\
Language instruction & 84.62 & 98.59 & 68.78 & 87.99 & 84.45 & 1,537 \\
Lighting & 94.52 & 96.63 & 96.42 & 76.28 & 91.16 & 1,142 \\
Background texture & 81.40 & 76.21 & 44.84 & 38.41 & 59.11 & 1,076 \\
Sensor noise & 71.79 & 86.26 & 54.88 & 60.36 & 68.39 & 1,601 \\
Object layout & 78.18 & 80.65 & 51.76 & 75.32 & 70.89 & 1,525 \\
\textit{All perturbations} & 72.19 & 79.75 & 53.07 & 59.90 & 66.06 & 10,030 \\
\midrule
\multicolumn{7}{@{}l}{\textbf{16 LBQs}} \\
Camera viewpoint & 65.43 & 74.75 & 19.12 & 39.86 & 49.22 & 1,599 \\
Robot initial state & 49.14 & 58.79 & 34.72 & 58.78 & 50.26 & 1,550 \\
Language instruction & 79.49 & 93.50 & 64.88 & 77.55 & 78.33 & 1,537 \\
Lighting & 99.32 & 97.31 & 89.25 & 70.07 & 89.32 & 1,142 \\
Background texture & 77.52 & 81.85 & 54.09 & 32.87 & 60.41 & 1,076 \\
Sensor noise & 80.34 & 86.97 & 57.52 & 60.58 & 71.14 & 1,601 \\
Object layout & 82.08 & 80.89 & 61.41 & 80.13 & 75.61 & 1,525 \\
\textit{All perturbations} & 75.60 & 81.25 & 52.72 & 59.71 & 67.12 & 10,030 \\
\bottomrule
\end{tabular}
\end{table}

\clearpage
\begin{table}[H]
\centering
\caption{\textbf{LIBERO-Plus LBQ sequence-length results by suite and perturbation (2/2).} Success rates (\%) for 24, 36, 64 LBQs. Pooled rates use episode counts; $N$ counts episodes across suites.}
\label{tab:lbq_plus_suite_2}
\footnotesize
\setlength{\tabcolsep}{3pt}
\renewcommand{\arraystretch}{1.03}
\begin{tabular}{@{}lrrrrrr@{}}
\toprule
Perturbation & Spatial & Object & Goal & Long & Pooled & $N$ \\
\midrule
\multicolumn{7}{@{}l}{\textbf{24 LBQs}} \\
Camera viewpoint & 71.28 & 78.03 & 32.35 & 41.53 & 55.22 & 1,599 \\
Robot initial state & 56.00 & 51.26 & 47.92 & 67.68 & 55.61 & 1,550 \\
Language instruction & 90.51 & 98.59 & 71.22 & 88.51 & 86.73 & 1,537 \\
Lighting & 98.97 & 98.65 & 98.21 & 81.75 & 94.57 & 1,142 \\
Background texture & 85.27 & 83.47 & 51.60 & 46.02 & 65.52 & 1,076 \\
Sensor noise & 84.33 & 92.42 & 63.32 & 67.71 & 76.83 & 1,601 \\
Object layout & 88.05 & 83.37 & 60.94 & 82.69 & 78.16 & 1,525 \\
\textit{All perturbations} & 81.64 & 82.92 & 59.36 & 67.41 & 72.63 & 10,030 \\
\midrule
\multicolumn{7}{@{}l}{\textbf{36 LBQs}} \\
Camera viewpoint & 50.27 & 80.56 & 22.79 & 39.62 & 47.97 & 1,599 \\
Robot initial state & 51.43 & 62.56 & 44.01 & 64.89 & 55.74 & 1,550 \\
Language instruction & 83.85 & 99.15 & 74.15 & 89.82 & 86.27 & 1,537 \\
Lighting & 98.97 & 98.99 & 98.21 & 82.12 & 94.75 & 1,142 \\
Background texture & 79.07 & 88.31 & 72.95 & 43.25 & 69.98 & 1,076 \\
Sensor noise & 75.50 & 92.42 & 68.60 & 73.27 & 77.70 & 1,601 \\
Object layout & 84.68 & 81.39 & 63.29 & 85.90 & 78.10 & 1,525 \\
\textit{All perturbations} & 74.10 & 85.39 & 61.17 & 67.96 & 72.05 & 10,030 \\
\midrule
\multicolumn{7}{@{}l}{\textbf{64 LBQs}} \\
Camera viewpoint & 45.21 & 57.58 & 13.97 & 35.56 & 37.77 & 1,599 \\
Robot initial state & 45.43 & 62.06 & 36.43 & 61.58 & 51.42 & 1,550 \\
Language instruction & 74.87 & 96.61 & 60.73 & 86.16 & 78.92 & 1,537 \\
Lighting & 96.58 & 97.98 & 97.13 & 70.80 & 90.89 & 1,142 \\
Background texture & 82.17 & 72.58 & 52.31 & 46.02 & 62.45 & 1,076 \\
Sensor noise & 50.71 & 71.33 & 44.33 & 57.46 & 56.53 & 1,601 \\
Object layout & 70.39 & 80.89 & 56.47 & 80.45 & 71.34 & 1,525 \\
\textit{All perturbations} & 65.11 & 76.05 & 49.44 & 61.81 & 62.98 & 10,030 \\
\bottomrule
\end{tabular}
\end{table}

\clearpage
\begin{table}[H]
\centering
\caption{\textbf{LIBERO-Plus LBQ sequence-length results by difficulty (1/2).} Success rates (\%) computed by pooling episodes across suites for 1, 8, 16 LBQs. Unspec. contains 121 lighting episodes without difficulty annotations, retained in all pooled totals. -- denotes no episodes.}
\label{tab:lbq_plus_difficulty_1}
\footnotesize
\setlength{\tabcolsep}{3pt}
\renewcommand{\arraystretch}{1.03}
\begin{tabular}{@{}lrrrrrrr@{}}
\toprule
Perturbation & Level 1 & Level 2 & Level 3 & Level 4 & Level 5 & Unspec. & Pooled \\
\midrule
\multicolumn{8}{@{}l}{\textbf{1 LBQ}} \\
Camera viewpoint & 68.34 & 43.33 & 31.69 & 37.72 & 27.52 & -- & 38.46 \\
Robot initial state & 83.87 & 78.77 & 61.50 & 48.77 & 31.91 & -- & 59.29 \\
Language instruction & 90.24 & 84.64 & 88.80 & 85.37 & 71.93 & -- & 83.86 \\
Lighting & 93.69 & 92.39 & 87.70 & 82.64 & 74.01 & 87.60 & 85.29 \\
Background texture & 72.53 & 62.36 & 47.53 & 33.74 & 10.96 & -- & 49.63 \\
Sensor noise & 82.49 & 82.80 & 79.67 & 59.53 & 35.68 & -- & 69.52 \\
Object layout & 81.90 & 79.14 & 78.72 & 70.43 & 51.21 & -- & 72.92 \\
\textit{All perturbations} & 81.75 & 73.89 & 66.86 & 61.98 & 43.69 & 87.60 & 65.36 \\
\midrule
\multicolumn{8}{@{}l}{\textbf{8 LBQs}} \\
Camera viewpoint & 62.31 & 53.06 & 49.54 & 39.04 & 27.93 & -- & 43.84 \\
Robot initial state & 81.85 & 69.18 & 51.07 & 35.79 & 21.65 & -- & 49.94 \\
Language instruction & 90.85 & 90.36 & 91.51 & 86.28 & 66.37 & -- & 84.45 \\
Lighting & 99.10 & 94.92 & 91.80 & 91.32 & 81.06 & 95.04 & 91.16 \\
Background texture & 79.83 & 70.11 & 58.17 & 45.40 & 22.60 & -- & 59.11 \\
Sensor noise & 79.66 & 77.40 & 77.00 & 62.40 & 40.25 & -- & 68.39 \\
Object layout & 80.17 & 78.81 & 74.16 & 69.65 & 48.79 & -- & 70.89 \\
\textit{All perturbations} & 81.57 & 74.84 & 68.82 & 62.94 & 42.92 & 95.04 & 66.06 \\
\midrule
\multicolumn{8}{@{}l}{\textbf{16 LBQs}} \\
Camera viewpoint & 73.87 & 58.33 & 48.92 & 43.86 & 35.11 & -- & 49.22 \\
Robot initial state & 79.84 & 70.89 & 53.74 & 37.54 & 18.80 & -- & 50.26 \\
Language instruction & 89.63 & 85.36 & 85.71 & 75.91 & 58.48 & -- & 78.33 \\
Lighting & 93.69 & 93.40 & 93.03 & 85.95 & 80.18 & 95.04 & 89.32 \\
Background texture & 78.97 & 70.11 & 60.84 & 45.40 & 28.77 & -- & 60.41 \\
Sensor noise & 81.36 & 84.20 & 79.33 & 63.97 & 37.76 & -- & 71.14 \\
Object layout & 86.49 & 85.10 & 79.94 & 69.26 & 53.29 & -- & 75.61 \\
\textit{All perturbations} & 83.45 & 77.57 & 70.20 & 61.56 & 43.49 & 95.04 & 67.12 \\
\bottomrule
\end{tabular}
\end{table}

\clearpage
\begin{table}[H]
\centering
\caption{\textbf{LIBERO-Plus LBQ sequence-length results by difficulty (2/2).} Success rates (\%) computed by pooling episodes across suites for 24, 36, 64 LBQs. Unspec. contains 121 lighting episodes without difficulty annotations, retained in all pooled totals. -- denotes no episodes.}
\label{tab:lbq_plus_difficulty_2}
\footnotesize
\setlength{\tabcolsep}{3pt}
\renewcommand{\arraystretch}{1.03}
\begin{tabular}{@{}lrrrrrrr@{}}
\toprule
Perturbation & Level 1 & Level 2 & Level 3 & Level 4 & Level 5 & Unspec. & Pooled \\
\midrule
\multicolumn{8}{@{}l}{\textbf{24 LBQs}} \\
Camera viewpoint & 70.35 & 65.83 & 61.23 & 51.32 & 39.01 & -- & 55.22 \\
Robot initial state & 85.89 & 81.85 & 59.36 & 37.19 & 23.36 & -- & 55.61 \\
Language instruction & 95.12 & 88.93 & 88.42 & 87.50 & 74.85 & -- & 86.73 \\
Lighting & 99.10 & 97.46 & 96.72 & 95.04 & 86.34 & 95.87 & 94.57 \\
Background texture & 83.69 & 73.80 & 65.02 & 52.76 & 36.30 & -- & 65.52 \\
Sensor noise & 83.62 & 86.20 & 81.67 & 74.93 & 49.38 & -- & 76.83 \\
Object layout & 87.07 & 86.75 & 85.11 & 75.49 & 52.94 & -- & 78.16 \\
\textit{All perturbations} & 86.44 & 82.20 & 75.55 & 69.30 & 50.36 & 95.87 & 72.63 \\
\midrule
\multicolumn{8}{@{}l}{\textbf{36 LBQs}} \\
Camera viewpoint & 65.33 & 50.83 & 47.69 & 48.68 & 38.60 & -- & 47.97 \\
Robot initial state & 84.68 & 78.77 & 58.29 & 42.46 & 24.22 & -- & 55.74 \\
Language instruction & 94.21 & 92.50 & 91.89 & 85.37 & 70.18 & -- & 86.27 \\
Lighting & 97.30 & 98.98 & 99.18 & 93.39 & 84.14 & 99.17 & 94.75 \\
Background texture & 81.97 & 76.38 & 71.48 & 58.90 & 48.63 & -- & 69.98 \\
Sensor noise & 85.88 & 87.60 & 80.67 & 75.72 & 50.62 & -- & 77.70 \\
Object layout & 90.52 & 88.41 & 83.28 & 70.82 & 52.94 & -- & 78.10 \\
\textit{All perturbations} & 86.07 & 80.79 & 74.36 & 69.25 & 50.41 & 99.17 & 72.05 \\
\midrule
\multicolumn{8}{@{}l}{\textbf{64 LBQs}} \\
Camera viewpoint & 62.81 & 43.89 & 32.31 & 32.89 & 28.95 & -- & 37.77 \\
Robot initial state & 79.44 & 71.58 & 53.48 & 40.70 & 21.37 & -- & 51.42 \\
Language instruction & 84.76 & 85.71 & 88.42 & 76.52 & 62.87 & -- & 78.92 \\
Lighting & 93.69 & 92.89 & 90.16 & 92.56 & 82.82 & 98.35 & 90.89 \\
Background texture & 82.83 & 69.74 & 56.65 & 50.92 & 39.73 & -- & 62.45 \\
Sensor noise & 70.62 & 73.20 & 59.00 & 42.82 & 30.29 & -- & 56.53 \\
Object layout & 84.48 & 82.78 & 74.16 & 63.04 & 47.75 & -- & 71.34 \\
\textit{All perturbations} & 80.05 & 72.43 & 63.23 & 57.00 & 42.63 & 98.35 & 62.98 \\
\bottomrule
\end{tabular}
\end{table}

\clearpage
\paragraph{IDM video-generation details.}
Tables~\ref{tab:idm_video_full_suite} and~\ref{tab:idm_video_full_category}
expand the two models in Table~\ref{tab:idm_video_generation} by suite and
perturbation category. Both models are evaluated on the same 1,415,986 action-replay reference
clips from 10,030 LIBERO-Plus episodes. All five metrics retain the complete
nine-frame scope, including the reconstructed conditioning frame.
Each valid clip is assigned the same weight in the evaluation; overlapping clips are correlated, and
longer episodes contribute more clips. FID and FVD use pooled feature
statistics rather than averages of group scores.

\begin{table}[H]
\centering
\caption{\textbf{IDM video-generation quality by LIBERO-Plus suite.} Detailed results for Table~\ref{tab:idm_video_generation}, using complete nine-frame clips including the reconstructed conditioning frame. Bold identifies the better score obtained by the two models within each group. The overall result uses all of the 1,415,986 clips; FID and FVD are computed from pooled features, not averaged across suites.}
\label{tab:idm_video_full_suite}
\footnotesize
\setlength{\tabcolsep}{3pt}
\renewcommand{\arraystretch}{1.03}
\begin{tabular}{@{}lrrrrr@{}}
\toprule
Method & FID $\downarrow$ & FVD$_9$ $\downarrow$ & SSIM $\uparrow$ & LPIPS $\downarrow$ & PSNR (dB) $\uparrow$ \\
\midrule
\multicolumn{6}{@{}l}{\textbf{Spatial} (224,794 clips)} \\
Fast-WAM IDM & 21.2074 & 55.7766 & 0.777271 & 0.195502 & 20.9940 \\
\sysname{} IDM & \textbf{21.0625} & \textbf{52.6436} & \textbf{0.779674} & \textbf{0.193154} & \textbf{21.0241} \\
\midrule
\multicolumn{6}{@{}l}{\textbf{Object} (293,546 clips)} \\
Fast-WAM IDM & 15.7660 & 70.1651 & 0.744659 & 0.204595 & 20.0774 \\
\sysname{} IDM & \textbf{15.5139} & \textbf{69.4985} & \textbf{0.746261} & \textbf{0.201890} & \textbf{20.1057} \\
\midrule
\multicolumn{6}{@{}l}{\textbf{Goal} (283,895 clips)} \\
Fast-WAM IDM & 20.7154 & 58.5497 & 0.753241 & 0.225478 & 19.9409 \\
\sysname{} IDM & \textbf{20.1850} & \textbf{54.0461} & \textbf{0.755572} & \textbf{0.221052} & \textbf{19.9956} \\
\midrule
\multicolumn{6}{@{}l}{\textbf{Long} (613,751 clips)} \\
Fast-WAM IDM & 12.8063 & \textbf{44.9536} & 0.790783 & 0.194223 & 20.7859 \\
\sysname{} IDM & \textbf{12.5229} & 45.1368 & \textbf{0.793103} & \textbf{0.191731} & \textbf{20.8387} \\
\midrule
\multicolumn{6}{@{}l}{\textbf{Overall} (1,415,986 clips)} \\
Fast-WAM IDM & 11.3573 & 30.8644 & 0.771549 & 0.202842 & 20.5026 \\
\sysname{} IDM & \textbf{11.2093} & \textbf{30.7990} & \textbf{0.773735} & \textbf{0.199942} & \textbf{20.5471} \\
\bottomrule
\end{tabular}
\end{table}

\clearpage
\begin{table}[H]
\centering
\caption{\textbf{IDM video-generation quality by LIBERO-Plus perturbation.} The same two models and complete nine-frame evaluation as Table~\ref{tab:idm_video_generation}. Bold marks the better score per perturbation group.}
\label{tab:idm_video_full_category}
\footnotesize
\setlength{\tabcolsep}{3pt}
\renewcommand{\arraystretch}{1.03}
\begin{tabular}{@{}lrrrrr@{}}
\toprule
Method & FID $\downarrow$ & FVD$_9$ $\downarrow$ & SSIM $\uparrow$ & LPIPS $\downarrow$ & PSNR (dB) $\uparrow$ \\
\midrule
\multicolumn{6}{@{}l}{\textbf{Background texture} (157,180 clips)} \\
Fast-WAM IDM & 20.7995 & 82.9152 & 0.712785 & 0.229659 & \textbf{19.6438} \\
\sysname{} IDM & \textbf{19.9457} & \textbf{73.7302} & \textbf{0.713475} & \textbf{0.229351} & 19.6018 \\
\midrule
\multicolumn{6}{@{}l}{\textbf{Camera viewpoint} (225,967 clips)} \\
Fast-WAM IDM & 15.7026 & 73.4651 & 0.759493 & 0.204473 & 20.0905 \\
\sysname{} IDM & \textbf{15.1467} & \textbf{73.1714} & \textbf{0.762843} & \textbf{0.200818} & \textbf{20.1603} \\
\midrule
\multicolumn{6}{@{}l}{\textbf{Language instruction} (216,029 clips)} \\
Fast-WAM IDM & \textbf{17.4338} & 42.8159 & 0.793574 & 0.166955 & 20.8744 \\
\sysname{} IDM & 17.5053 & \textbf{42.5802} & \textbf{0.794395} & \textbf{0.166394} & \textbf{20.9292} \\
\midrule
\multicolumn{6}{@{}l}{\textbf{Lighting} (160,483 clips)} \\
Fast-WAM IDM & 22.8356 & 83.6936 & 0.792433 & 0.184414 & 21.9082 \\
\sysname{} IDM & \textbf{22.6035} & \textbf{81.1291} & \textbf{0.796552} & \textbf{0.181510} & \textbf{22.0167} \\
\midrule
\multicolumn{6}{@{}l}{\textbf{Object layout} (204,567 clips)} \\
Fast-WAM IDM & \textbf{9.3753} & \textbf{25.0171} & 0.784911 & \textbf{0.177259} & 20.5293 \\
\sysname{} IDM & 9.4114 & 26.0252 & \textbf{0.785688} & 0.177485 & \textbf{20.5472} \\
\midrule
\multicolumn{6}{@{}l}{\textbf{Robot initial state} (217,599 clips)} \\
Fast-WAM IDM & 10.3116 & \textbf{23.8363} & 0.778736 & \textbf{0.199732} & 19.9008 \\
\sysname{} IDM & \textbf{10.2443} & 24.3979 & \textbf{0.779363} & 0.200246 & \textbf{19.9220} \\
\midrule
\multicolumn{6}{@{}l}{\textbf{Sensor noise} (234,161 clips)} \\
Fast-WAM IDM & 33.3411 & 110.3438 & 0.769643 & 0.254248 & 20.7066 \\
\sysname{} IDM & \textbf{31.6984} & \textbf{106.0439} & \textbf{0.774328} & \textbf{0.242273} & \textbf{20.7761} \\
\bottomrule
\end{tabular}
\end{table}

\clearpage
\subsection{Latency Measurement Details}
\label{sec:appendix_efficiency}

\paragraph{Simulation profiling on NVIDIA H20Z.}
The LIBERO efficiency results reported in Table~\ref{tab:inference_efficiency} are
measured for \sysname{}, Fast-WAM, and Faster-WAM on NVIDIA H20Z GPUs in
\texttt{bfloat16}, using an identical cached LIBERO input, an action horizon
of 32, and 10 denoising steps. We average 10 measured inference calls after
three warm-up calls. Timing begins immediately before Video DiT layer~0 and
ends at the final clean action latent, excluding observation preprocessing,
VAE encoding, text encoding, and postprocessing. Under this core-inference
protocol, \sysname{} takes $88.05$\,ms per call, compared with $405.91$\,ms for
Fast-WAM, yielding a $4.61\times$ speedup. Runtime-shape FLOPs count linear
and attention matrix multiplications and linear bias additions, with one
multiply--add counted as two FLOPs; action FLOPs include conditioning and all
denoising steps. These FLOP counts explicitly exclude normalization, activations, RoPE,
softmax, and other elementwise operations, including video K/V fusion.\looseness=-1

The Fast-WAM and \sysname{} values are retained from earlier profiling runs.
Faster-WAM was measured separately on an otherwise idle GPU, with a
CUDA-event latency of $321.16$\,ms; \sysname{} achieves a $3.65\times$
speedup over Faster-WAM under this core-inference protocol.
Its native one-pass future-cache configuration processes nine video frames,
encoded as three latent frames with 294 video tokens, whereas the other two
models process one observed latent frame with 98 video tokens. The cached
observation and text inputs are identical across all three models. Faster-WAM
uses 30 video and 30 action layers, with $2.745$\,T video FLOPs and
$0.476$\,T action FLOPs ($3.221$\,T in total). Peak memory denotes per-process
allocated GPU memory, including loaded parameters, in MiB ($2^{20}$ bytes),
rather than the total amount of memory used on the device.

\paragraph{Real-robot inference on NVIDIA RTX 5090.}
For the real-robot experiments on the \textit{PiPER X from SongLing}, we perform the model's inference on an NVIDIA GeForce RTX 5090 GPU. In this setting,
\sysname{} generates an action chunk in $77.1$\,ms, as also reported in
Fig.~\ref{fig:real_world_results}.

\end{document}